\documentclass[10pt,twocolumn,letterpaper]{article}

\usepackage[pagenumbers]{cvpr} 
\usepackage{booktabs} 
\usepackage{multirow}
\usepackage{array}
\usepackage{bbding}
\usepackage{makecell}

\usepackage{marvosym}
\usepackage{comment}

\newcommand{\ours}{OmniVGGT\xspace}

\definecolor{cvprblue}{rgb}{0.21,0.49,0.74}
\usepackage[pagebackref,breaklinks,colorlinks,allcolors=cvprblue]{hyperref}

\def\paperID{11131} 
\def\confName{CVPR}
\def\confYear{2026}

\title{\emph{\textcolor[RGB]{38, 89, 146}{Omni}\textcolor[RGB]{199, 64, 52}{VGGT}}: Omni-Modality Driven Visual Geometry Grounded Transformer}

\author{
Haosong Peng$^{1}$$^\ast$,  
Hao Li$^{2}$$^\ast$,
Yalun Dai$^{2}$,
Yushi Lan$^{2}$,
Yihang Luo$^{2}$,
Tianyu Qi$^{3}$,
\\
Zhengshen Zhang $^{4}$,
Yufeng Zhan$^{1}$\textsuperscript{\Letter},
Junfei Zhang$^{5}$\textsuperscript{\Letter},
Wenchao Xu$^{1}$\textsuperscript{\Letter},
Ziwei Liu$^2$    
\\
$^1$HKUST \quad
$^2$NTU \quad
$^3$SYSU \quad
$^4$NUS \quad
$^5$Alibaba Group
}

\begin{document}
 \twocolumn[{%
 \renewcommand\twocolumn[1][]{#1}%
 \maketitle
 \begin{center}
 \vspace{-10pt}
 \label{fig:teaser}
 \centering\includegraphics[width=1\textwidth]{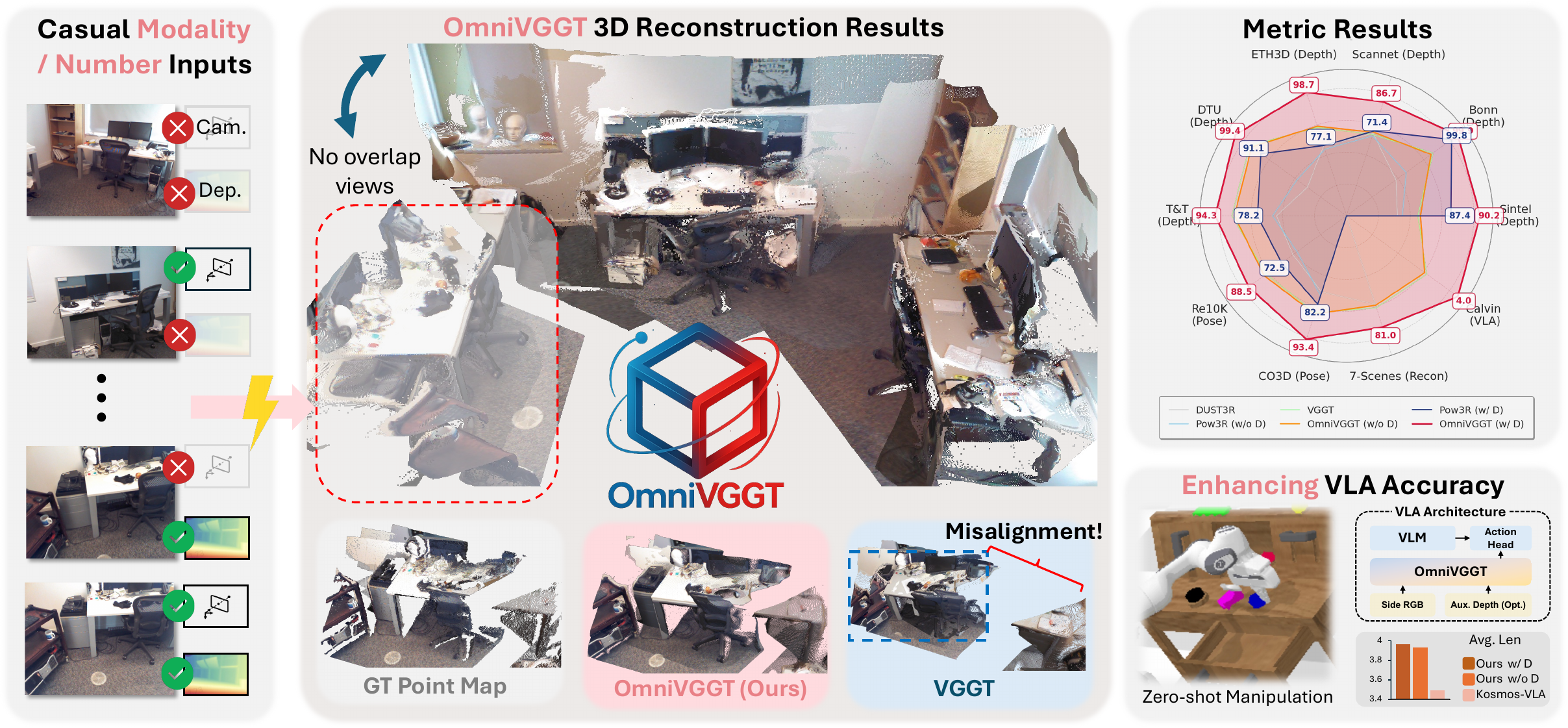} 
 \captionof{figure}{
 We proposed \ours, a spatial foundation model that can effectively benefit from an arbitrary number of auxiliary geometric
modalities (depth, camera intrinsics \& pose) to obtain high-quality 3D geometric results. Experimental results show that OmniVGGT
achieves state-of-the-art performance across various downstream tasks and further improves performance on robot manipulation tasks.
 }
 \end{center}%
 }]
 
{
  \renewcommand{\thefootnote}%
    {}
  \footnotetext[2]{ $^\ast$Equal Contribution. \textsuperscript{\Letter}Corresponding Authors.}
}

\begin{abstract}
General 3D foundation models have started to lead the trend of unifying diverse vision tasks, yet most assume RGB-only inputs and ignore readily available geometric cues (e.g., camera intrinsics, poses, and depth maps). 
To address this issue, we introduce \textbf{\ours}, a novel framework that can effectively benefit from an arbitrary number of auxiliary geometric modalities during both training and inference.
In our framework, a \textbf{GeoAdapter} is proposed to encode depth and camera intrinsics/extrinsics into a spatial foundation model.
It employs zero-initialized convolutions to progressively inject geometric information without disrupting the foundation model's representation space. 
This design ensures stable optimization with negligible overhead, maintaining inference speed comparable to VGGT even with multiple additional inputs.
Additionally, a \textbf{stochastic multimodal fusion} regimen is proposed, which randomly samples modality subsets per instance during training. 
This enables an arbitrary number of modality inputs during testing and promotes learning robust spatial representations instead of overfitting to auxiliary cues.
Extensive experiments on monocular/multi-view depth estimation, multi-view stereo, and camera pose estimation demonstrate that \ours outperforms prior methods with auxiliary inputs and achieves state-of-the-art results even with RGB-only input. 
To further highlight its practical utility, we integrated \ours into \textbf{vision-language-action (VLA) models}. 
The enhanced VLA model by \ours not only outperforms the vanilla point-cloud-based baseline on mainstream benchmarks, but also effectively leverages accessible auxiliary inputs to achieve consistent gains on robotic tasks.
Project Page: \url{https://livioni.github.io/OmniVGGT-official/} 
\end{abstract}

\section{Introduction}
\label{sec:intro}

Similar to how large language models (LLMs, e.g., ChatGPT \citep{achiam2023gpt}) have revolutionized the field of NLP, building a general model for 3D perception (e.g., VGGT \citep{wang2025vggt}) is emerging as a new paradigm in the 3D vision community. 
Unlike traditional approaches that focus on solving specific tasks (e.g., monocular and stereo depth estimation \citep{li2018megadepth,hu2025depthcrafter}, pose estimation \citep{wang2023posediffusion,arnold2022map}, or novel view synthesis \citep{mildenhall2021nerf,kerbl20233dgs}), recent methods \citep{wang2024dust3r,leroy2024mast3r,wang2025vggt} have demonstrated tremendous potential in addressing all these tasks in a unified manner using feed-forward architectures.

While these general solutions aim to establish unified representations for various 3D downstream tasks, they typically overlook efforts to \textit{unify diverse input modalities}. 
Most methods~\citep{wang2024dust3r,leroy2024mast3r,wang2025vggt} are limited to accepting only RGB images as input, neglecting the rich multimodal inputs commonly found in 3D vision, such as depth maps and camera poses.
Recent Pow3R~\citep{jang2025pow3r} has made attempts to incorporate multimodal inputs, but they are typically limited to handling at most two inputs (e.g., RGB pair and depth pair). 
However, in real-world 3D vision applications, an \textit{arbitrary number} of multimodal inputs is often available. 
For example, virtual/augmented reality (VR/AR) utilizes RGB-D data to acquire depth maps~\citep{Dai2017BundleFusion,Hu2024CGSLAM,Peng2024RTGSLAM}, some autonomous driving strategies leverage LiDAR to capture point clouds~\citep{Sun2020WaymoOpenDataset,Caesar2020nuScenes,Lang2019PointPillars}, and robotic applications may incorporate knowledge of camera intrinsics and/or extrinsics (e.g., pose)~\citep{hu2023towardrb,keetha2024splatamrb}. 
Therefore, the ability to \textit{seamlessly benefit from these accessible multimodal inputs to enhance model performance} is of critical importance.

To address this problem, we propose \ours, which introduces a flexible input scheme that can benefit from an arbitrary number of various geometric modalities when available. 
First, we introduce a lightweight \textbf{GeoAdapter} to effectively incorporate geometric cues (depth and camera parameters). 
A key challenge arises from their differing properties: unlike a depth map, which provides dense, per-pixel spatial cues, the camera pose is a global attribute.
Therefore, directly injecting encoded 3D camera information can destabilize the feature space of large-scale foundation models, leading to severe issues during early training.
To mitigate this and preserve the foundation model's high-quality feature representations, we employ zero convolution to process the camera pose, progressively initializing the adapter's parameters from zero.
This strategy ensures training stability and maintains the foundation model's superior representations, with negligible added computational overhead. Our lightweight design ensures that despite multiple additional inputs, the inference speed of \ours remains comparable to that of the vanilla VGGT.

Second, powered by our \textbf{stochastic multimodal fusion strategy}, \ours can leverage arbitrary numbers of geometric modality inputs during testing, significantly enhancing its practicality and generality compared to vanilla VGGT and Pow3R. 
Notably, this stochastic strategy enables the model to learn more generalized and robust 3D spatial representations from diverse auxiliary modalities, rather than merely fitting the additional 3D information. 

Extensive experiments on various 3D vision tasks (e.g., mono-/multi-view depth estimation, multi-view stereo, and camera pose estimation) demonstrate that our method not only significantly outperforms existing approaches when auxiliary modality inputs are available but also surpasses state-of-the-art methods when using RGB-only inputs.
Moreover, to thoroughly validate the practical value of our model, we integrate \ours with vision-language-action (VLA) models and conduct extensive robotic manipulation experiments. 
Evaluations show that the enhanced VLA model exhibits superior spatial understanding capabilities compared to the vanilla point-cloud-based baseline, while simultaneously demonstrating its capacity to exploit accessible auxiliary inputs (e.g., depth) for consistent gains on robotic tasks.

\section{Related Work}

\begin{figure*}[!t]
\begin{center}
\includegraphics[width=\linewidth]{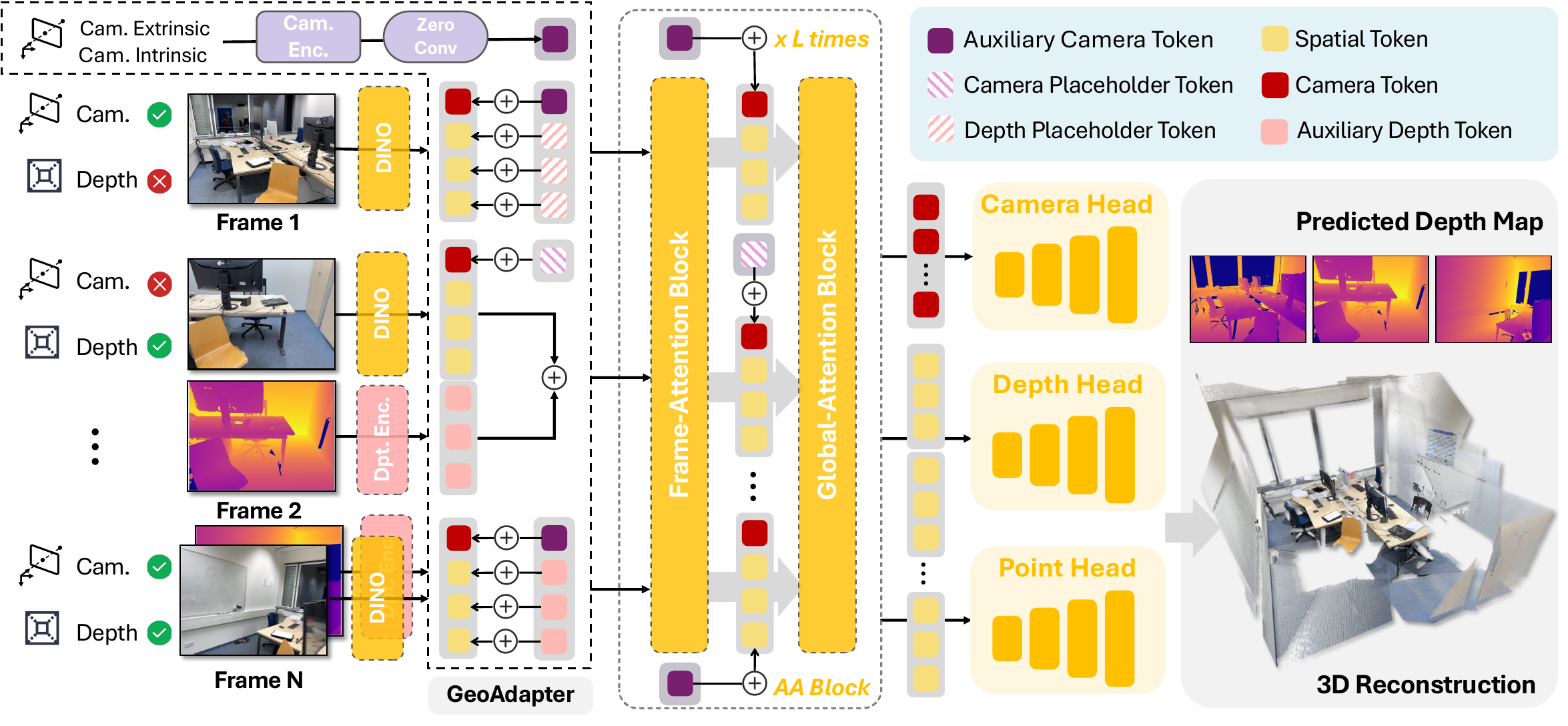}
\caption{\textbf{Overview of \ours.} \ours takes as input a set of images together with an arbitrary number of corresponding camera parameters (poses and intrinsics) or depth maps. 
Camera placeholder tokens and depth placeholder tokens are used to substitute the tokens for which auxiliary information is missing.
The inputs are processed through \(L\) layers of Alternating-Attention, and finally, three prediction heads are employed to output depth maps, camera poses, and 3D point maps.}
\label{fig:overview}
\end{center}
\vspace{-7mm}
\end{figure*}

\subsection{3D Reconstruction}
Traditional methods \citep{yao2018mvsnet,huang2018deepmvs,yao2020blendedmvs,cao2024mvsformer++} employed a Structure-from-Motion (SfM) framework to ascertain camera positions and produce sparse point clouds. 
Recently, advancements such as Neural Radiance Fields (NeRF) \citep{mildenhall2021nerf} and 3D Gaussian Splatting (3DGS) \citep{kerbl20233dgs} have become ubiquitous in 3D vision thanks to their photo-realistic characteristics and superior novel-view synthesis capabilities, enabling applications from enhancing
immersive environments in VR/AR and improving spatial awareness in robotics, to supporting urban planning and cultural heritage digitization \citep{gao2022nerf,wang2024nerfs,3dgssurvey2024,3dgssurvey2025}. 
However, these methods typically require per-scene optimization and accurate camera poses as input, which limits their generalization and scalability.

\subsection{Spatial Foundation Model}
More recently, advancements in deep learning have introduced novel alternatives to traditional SfM methods. 
DUSt3R \cite{wang2024dust3r} represents a significant deviation from conventional SfM pipelines by predicting point clouds from image pairs without relying on geometric constraints or inductive biases. 
Building on this paradigm, several works have proposed variations with distinct architectural innovations. MASt3R~\cite{leroy2024mast3r} improves the estimation of the pixel-wise correspondence between image pairs, strengthening the efficacy of unconstrained feed-forward models for SfM tasks. CUT3R~\cite{wang2025cut3r} introduces a recurrent formulation of DUSt3R, achieving computational efficiency at the expense of marginal accuracy degradation. More recently, VGGT~\cite{wang2025vggt} proposes a multi-view architecture that processes multiple images simultaneously, moving beyond pairwise processing to improve reconstruction consistency and robustness.
Owing to its strong generalization ability, it has already been applied in fields such as SLAM~\cite{maggio2025vggt} and 4D reconstruction~\cite{lan2025stream3r,zhuo2025streaming}, and novel view synthesis~\cite{liu2025vggt}.
However, while these methods aim to create unified 3D representations broadly applicable to downstream tasks, they are often limited to RGB-only inputs, neglecting the substantial benefits afforded by rich auxiliary 3D modalities.

\section{Preliminary}
We follow the recently proposed VGGT~\cite{wang2025vggt} method that directly infers all key 3D attributes of a scene within seconds.
The input image set $\mathbf{I}$ is first patchified into spatial tokens $\mathbf{e}_f$ via DINO backbone~\cite{oquab2023dinov2}.
Spatial tokens from all frames are concatenated with learnable camera tokens $\mathbf{e}_c$ and register tokens $\mathbf{e}_r$, and then processed jointly by the transformer encoder \(\mathcal{E}\):
\begin{equation}
(\hat{\mathbf{e}}_c,\hat{\mathbf{e}}_r,\hat{\mathbf{e}}_f)=\mathcal{E}(\mathbf{e}_c,\mathbf{e}_r,\mathbf{e}_f).
\end{equation}
VGGT adopts an Alternating-Attention (AA) scheme as its encoder, where frame-wise self-attention captures intra-image structures, and global self-attention aggregates information across views.
After being processed by AA blocks ($L$ layers in total), refined spatial tokens $\hat{\mathbf{e}}_f$ and camera tokens $\hat{\mathbf{e}}_c$ are retained for prediction, while register tokens $\hat{\mathbf{e}}_r$ are discarded. 
For depth and point map predictions, VGGT uses the DPT heads~\cite{ranftl2021vision} as dense prediction heads, and feeds the dense feature to get the depth maps $\hat{\mathbf{D}}$, 3D point maps $\hat{\mathbf{P}}$, and corresponding confidence map $\hat{\mathbf{Y}}$.
For the camera head, all camera tokens are passed through four self-attention layers 
followed by a linear layer to predict the camera intrinsics and extrinsics.
\section{Methodology}
This section details OmniVGGT: 
1) We articulate its core capability to leverage arbitrary auxiliary inputs in Section~\ref{sec:Definition}. 
2) We introduce the GeoAdapter as our key architecture for multimodal injection in Section~\ref{sec:omnivggt}. 
3) We discuss the stochastic training strategy pivotal for multimodal fusion Section~\ref{sec:training}.

\subsection{Overview of OmniVGGT}\label{sec:Definition}
As illustrated in Fig.~\ref{fig:overview}, our OmniVGGT ingests an image set $\mathbf{I} = \{I_i\}^N_{i=1}$, along with an arbitrary number of auxiliary inputs such as camera parameters $\mathbf{C} = \{C_j\}^Q_{j=1}$ and depth maps $\mathbf{D} = \{D_k,M_k\}^O_{k=1}$ (where $Q\leq N, O\leq N$), to guide more accurate and robust 3D scene reconstruction.
Each image $I\in\mathbb{R}^{3\times H\times W}$ may be associated with known camera parameters $C=\{K,G\}$, i.e., intrinsics $K\in \mathbb{R}^{3\times 3}$ and pose $G=[R \mid t]\in\mathbb{R}^{4\times4}$, or with a known depth map $D\in\mathbb{R}^{H\times W}$ accompanied by a corresponding mask $M\in\{0,1\}^{H\times W}$ for valid depth.
All images are simultaneously fed into the OmniVGGT network, which fully leverages the available camera parameters and depth maps (if provided) to produce the predicted 3D point maps along with complete camera poses and intrinsics, depth maps, and corresponding confidence maps in an end-to-end manner.

\subsection{GeoAdapter for Stochastic Multimodal Inputs}\label{sec:omnivggt}

Our lightweight GeoAdapter consists of a camera adapter and a depth adapter, both of which include auxiliary geometric information normalization, encoding, and injection steps.
It seamlessly incorporates prior information into the encoder, enabling more effective feature fusion.

\noindent \textbf{Camera Adapter.}
First, given a set of camera intrinsics $\{K_j\}^Q_{j=1}$ and poses $\{G_j\}^Q_{j=1}=[R \mid t]$, we first align the coordinate system origin with the first camera. 
We then compute the average distance between the remaining cameras and the origin, using it as a scale factor to normalize all camera poses.
The formulations are as follows:
\begin{equation}
s = \frac{1}{Q-1} \sum_{j=2}^{Q} \| t_j - t_1 \|_2,  \quad j=2,\dots,Q,
\end{equation}
\begin{equation}
G_{j}^\prime = G_{j}G_{1}^{-1};\  t'_1 = \mathbf{0}; \  t^\prime_i = \frac{t_i - t_1}{s}; \ j=2,\dots,Q.
\end{equation}
Second, the intrinsics and normalized poses are parameterized into a feature vector $\{K,G\}$ as $\mathbf{g} = \{\mathbf{q}, \mathbf{t}, \mathbf{f}\}$ follow~\cite{wang2024vggsfm}, where $\mathbf{q} \in \mathbb{R}^{4}$ denotes the rotation quaternion, $\mathbf{t} \in \mathbb{R}^{3} $ denotes the translation vector, and $\mathbf{f} \in \mathbb{R}^{2}$ denotes the field of view.
After that, $\mathbf{g}$ is fed into a dedicated camera encoder \(\mathcal{E}^{\text{cam}}_l(\cdot)\) before  \(l\)-th AA block to obtain the auxiliary camera tokens $\mathbf{e}_c^{\mathrm{aux}}$ :
\begin{equation}
\mathbf{e}_{c,j,l}^{\mathrm{aux}}=\mathcal{E}^{cam}_l(\mathbf{g}_j),j = 0,\ldots,Q,\ l=0,\ldots,L,
\end{equation}
For images without camera information, we use a zero vector $\mathbf{e}_c^{\mathrm{plh}}$ as the camera placeholder token to replace the missing auxiliary camera token.
Third, we pass all the auxiliary camera tokens through a zero-conv layer \(\mathcal{ZC}_l(\cdot)\), and then add the processed tokens to the camera tokens:
\begin{equation}
\mathbf{e}_{c,i,l}^\prime = \mathbf{e}_{c,i,l} + \mathcal{ZC}_l\left(m_i (\mathbf{e}^{\mathrm{aux}}_{c,i,l}) + (1-m_i)\mathbf{e}_{c}^{\mathrm{plh}}\right),
\end{equation}
where $m_i\in\{0,1\}$ is a binary indicator that equals 1 if the image has auxiliary camera parameters and 0 otherwise.

\noindent \textbf{Depth Adapter.}
First, auxiliary depth maps $\{D\}_{k=1}^{O}$ are first normalized on a per-batch basis. 
For each depth map $\{D_k\}_{k=1}^{O}$, we identify valid pixels using the corresponding mask. 
The valid depths are then normalized by the mean depth value computed over all valid pixels in the batch and further concatenated along the channel dimension:
\begin{equation}
X=[D ; M],\quad X\in\mathbb{R}^{2\times H\times W}.
\end{equation}
Second, $X$ is fed into a depth encoder \(\mathcal{E}^{dpt}(\cdot)\) to obtain the auxiliary depth token $\mathbf{e}_d^{\mathrm{aux}}$, which consists of a dedicated convolutional layer that tokenizes $X$ and aligns the dimension of spatial token:
\begin{equation}
\mathbf{e}^{\mathrm{aux}}_{d,k} = \mathcal{E}^{dpt}(X_k), \quad k=0,\ldots,O.
\end{equation}
For images without reference depth information, we use a separate depth placeholder token instead. 
Third, the auxiliary depth token and placeholder token are then directly added to the corresponding spatial token:
\begin{equation}
\mathbf{e}_{f,i}^\prime=\mathbf{e}_{f,i}+\left(n_i(\mathbf{e}_{d,i}^{\mathrm{aux}})+(1-n_i)\mathbf{e}_{d}^{\mathrm{plh}}\right)
\end{equation}
where $n_i\in\{0,1\}$ is a binary indicator that equals 1 if the image has auxiliary depth maps and 0 otherwise.
Through extensive ablation study, we observe that applying an additional zero-conv layer to the depth branch is redundant, as it disrupts the effective integration of depth information.

\begin{table*}[t]
\caption{\textbf{Impact of Auxiliary Information Injection.} 
We evaluate the performance of depth and pose estimation when varying the percentage of injected ground-truth information on the unseen Sintel dataset~\cite{butler2012naturalistic}. 
The absolute improvements over the baseline (\textit{w/o} aux. information) are highlighted in \textcolor{teal}{green}, and the best results are shown in \textbf{bold}.}
\label{tab:maintable}
\vspace{-1mm}
\centering
\resizebox{0.8\linewidth}{!}{
\begin{tabular}{@{}c|cc|cc|ccc@{}}
\toprule
\textbf{Method} & \multicolumn{2}{c|}{\textbf{Aux. information (\%)}} & \multicolumn{2}{c|}{\textbf{Depth}} & \multicolumn{3}{c}{\textbf{Camera}} \\ \cmidrule(l){2-8} 
\textbf{} & \multicolumn{1}{c|}{Depth} & Camera& \multicolumn{1}{c|}{Abs Rel$\downarrow$} & $\delta<1.25\uparrow$& \multicolumn{1}{c|}{RRA@$5^{\circ}$$\uparrow$} & \multicolumn{1}{c|}{RTA@ $5^{\circ}$$\uparrow$} & AUC@$30^{\circ}\uparrow$\\ \midrule
VGGT & \multicolumn{1}{c|}{\XSolidBrush} & \XSolidBrush & \multicolumn{1}{c|}{0.722} &  70.81& \multicolumn{1}{c|}{95.69} & \multicolumn{1}{c|}{53.92} &  70.55\\ 
\textbf{\ours} & \multicolumn{1}{c|}{\XSolidBrush} & \XSolidBrush & \multicolumn{1}{c|}{\textbf{0.558}} &  \textbf{71.46}& \multicolumn{1}{c|}{\textbf{96.15}} & \multicolumn{1}{c|}{\textbf{54.01}} &  \textbf{70.83}\\ \midrule
\multirow{10}{*}{\thead{\textbf{\ours +} \\ \textbf{aux. information}}}  & \multicolumn{1}{c|}{30} & \XSolidBrush & \multicolumn{1}{c|}{0.169 \textcolor{teal}{(+0.389)}} &  78.92 \textcolor{teal}{(+7.46)}& \multicolumn{1}{c|}{96.65 \textcolor{teal}{(+0.50)}} & \multicolumn{1}{c|}{54.15 \textcolor{teal}{(+0.14)}} &  71.43 \textcolor{teal}{(+0.60)}\\
 & \multicolumn{1}{c|}{50} & \XSolidBrush & \multicolumn{1}{c|}{0.150 \textcolor{teal}{(+0.408)}} &  80.93 \textcolor{teal}{(+9.47)}& \multicolumn{1}{c|}{96.65 \textcolor{teal}{(+0.50)}} & \multicolumn{1}{c|}{55.90 \textcolor{teal}{(+1.89)}} &  71.49 \textcolor{teal}{(+0.66)}\\
 & \multicolumn{1}{c|}{70} & \XSolidBrush & \multicolumn{1}{c|}{0.124 \textcolor{teal}{(+0.434)}} &  83.24 \textcolor{teal}{(+11.78)}& \multicolumn{1}{c|}{96.65 \textcolor{teal}{(+0.50)}} & \multicolumn{1}{c|}{56.89 \textcolor{teal}{(+2.88)}} &  73.68 \textcolor{teal}{(+2.85)}\\
 & \multicolumn{1}{c|}{100} & \XSolidBrush & \multicolumn{1}{c|}{\textbf{0.106 \textcolor{teal}{(+0.452)}}} &  \textbf{85.95 \textcolor{teal}{(+14.49)}}& \multicolumn{1}{c|}{\textbf{96.93 \textcolor{teal}{(+0.78)}}} & \multicolumn{1}{c|}{\textbf{59.73 \textcolor{teal}{(+5.72)}}} &  \textbf{77.16 \textcolor{teal}{(+6.33)}}\\ \cmidrule(l){2-8}
 & \multicolumn{1}{c|}{\XSolidBrush} & 30 & \multicolumn{1}{c|}{0.555\textcolor{teal}{(+0.003)}} &  72.45 \textcolor{teal}{(+0.99)}& \multicolumn{1}{c|}{98.56 \textcolor{teal}{(+2.41)}} & \multicolumn{1}{c|}{61.44 \textcolor{teal}{(+7.43)}} &  75.87 \textcolor{teal}{(+5.04)}\\
 & \multicolumn{1}{c|}{\XSolidBrush} & 50& \multicolumn{1}{c|}{0.554 \textcolor{teal}{(+0.004)}} &  72.43 \textcolor{teal}{(+0.97)}& \multicolumn{1}{c|}{98.67 \textcolor{teal}{(+2.52)}} & \multicolumn{1}{c|}{63.76 \textcolor{teal}{(+9.75)}} &  77.90 \textcolor{teal}{(+7.07)}\\
 & \multicolumn{1}{c|}{\XSolidBrush} & 70& \multicolumn{1}{c|}{0.554 \textcolor{teal}{(+0.004)}} &  72.49 \textcolor{teal}{(+0.90)}& \multicolumn{1}{c|}{98.84 \textcolor{teal}{(+3.82)}} & \multicolumn{1}{c|}{66.91 \textcolor{teal}{(+12.9)}} &  79.72 \textcolor{teal}{(+8.80)}\\
 & \multicolumn{1}{c|}{\XSolidBrush} & 100 & \multicolumn{1}{c|}{\textbf{0.553 \textcolor{teal}{(+0.005)}}} & \textbf{72.36 \textcolor{teal}{(+0.90)}}& \multicolumn{1}{c|}{\textbf{99.97 \textcolor{teal}{(+3.82)}}} & \multicolumn{1}{c|}{\textbf{75.83 \textcolor{teal}{(+21.82)}}} &  \textbf{85.35 \textcolor{teal}{(+14.52)}}\\ \cmidrule(l){2-8}
 & \multicolumn{1}{c|}{100} & 100 & \multicolumn{1}{c|}{\textbf{0.106 \textcolor{teal}{(+0.452)}}} &  \textbf{85.95 \textcolor{teal}{(+14.49)}}& \multicolumn{1}{c|}{\textbf{99.97 \textcolor{teal}{(+3.82)}}} & \multicolumn{1}{c|}{\textbf{76.33 \textcolor{teal}{(+22.32)}}} &  \textbf{85.99 \textcolor{teal}{(+15.16)}}\\ \bottomrule
\end{tabular}}
\end{table*}

\subsection{Training Procedure}\label{sec:training}

\noindent\textbf{Training Objective.}
Following VGGT~\cite{wang2025vggt}, our training objective is a multi-task loss comprising three components: camera, depth, and point map supervision. The total loss is defined as $\mathcal{L} = \mathcal{L}_{\text{camera}} + \mathcal{L}_{\text{depth}} + \mathcal{L}_{\text{pmap}}$.
The camera loss, $\mathcal{L}_{\text{camera}}$, supervises the predicted camera parameters $\hat{\mathbf{g}}_i$ against the ground truth $\mathbf{g}_i$ using an $\ell_1$ regression loss. 
For the depth loss $\mathcal{L}_{\text{depth}}$ and the point map loss $\mathcal{L}_{\text{pmap}}$, we adopt the confidence-aware regression loss. Each of these is further augmented with a gradient-based term to enhance local geometric consistency.

\noindent \textbf{Stochastic Multimodal Fusion Strategy.}
To enable the ability that accepting arbitrary input combinations at inference time, we train the network on batches of images where auxiliary information is stochastically assigned.
Specifically, for an image sequence of length $S$, we first uniformly sample a number $Q \in [0, S]$ to determine the quantity of images to be provided with ground truth (GT) camera parameters. We then assign these camera annotations to the first $Q$ images in the sequence.
In contrast, for depth GT, a number $O \in [0, S]$ is sampled independently, and these $O$ annotations are assigned to randomly selected indices within the sequence.
This stochastic assignment strategy simplifies the training process while ensuring the model is robust to various partial-information scenarios during inference.
Furthermore, to ensure stability and handle the auxiliary-free case, a subset of training batches (with probability $p\%$) is processed using only raw RGB images.

\section{Experiments}

\begin{figure*}[t]
\begin{center}
\includegraphics[width=0.9\linewidth]{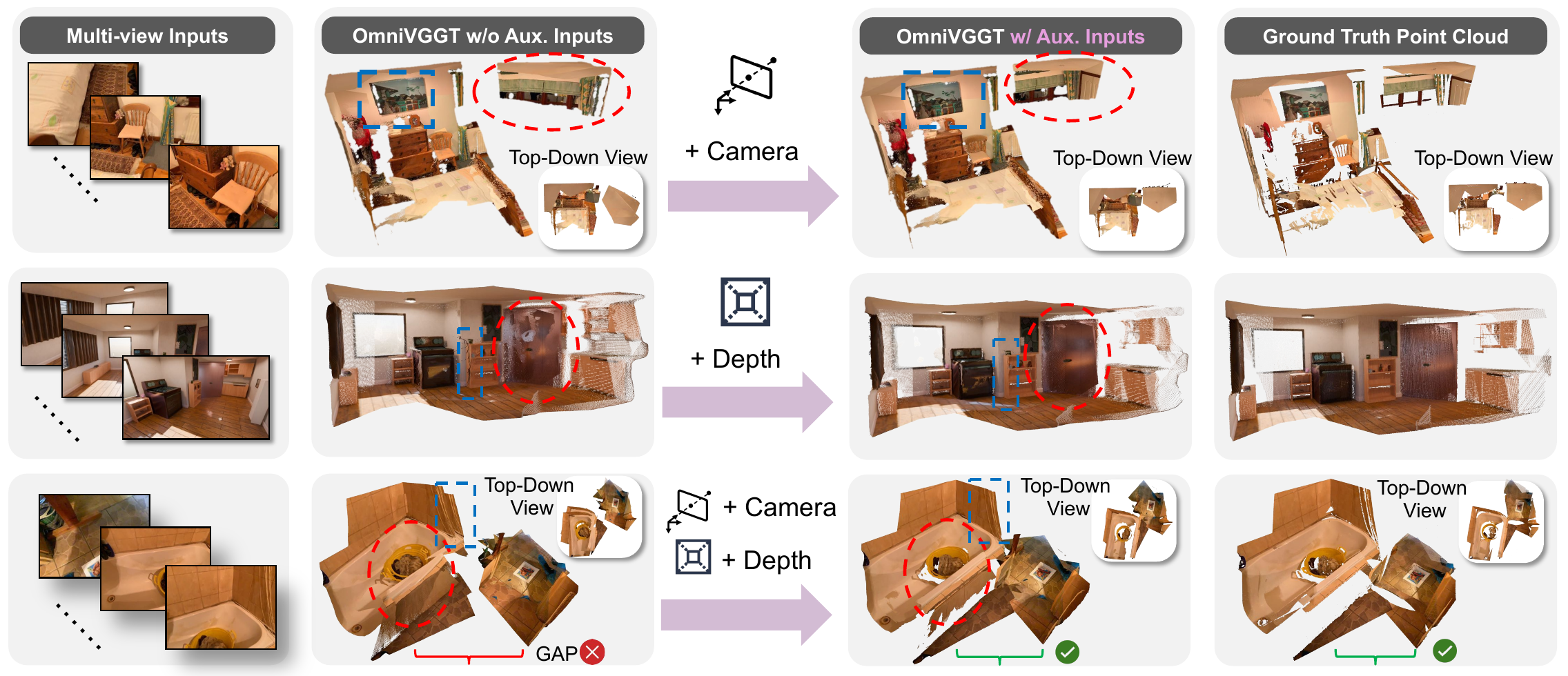}
\caption{\textbf{Visual Results of \ours with Different Auxiliary Information.} (Top) Camera information help correct challenging scenarios with little or no overlap. (Middle) Providing depth information leads to more accurate local geometry, such as on door surfaces. (Bottom) When both depth and camera information are provided, the relative distances and viewing angles are properly corrected.}
\label{fig:ablation}
\end{center}
\vspace{-8mm}
\end{figure*}

\textbf{Datasets.} We train our model using images from 19 public datasets, including: ARKitScenes~\cite{baruch2021arkitscenes}, BlendedMVS~\cite{yao2020blendedmvs}, DL3DV~\cite{ling2024dl3dv}, Dynamic Replica~\cite{karaev2023dynamicstereo}, HyperSim~\cite{roberts2021hypersim}, Kubric~\cite{greff2022kubric}, MapFree~\cite{arnold2022map}, MegaDepth~\cite{li2018megadepth}, Matterport 3D~\cite{ramakrishnan2021hm3d}, MVS-Synth~\cite{huang2018deepmvs}, ScanNet~\cite{dai2017scannet}, ScanNet++~\cite{yeshwanth2023scannet++}, Spring~\cite{mehl2023spring}, TartanAir~\cite{wang2020tartanair}, UASOL~\cite{bauer2019uasol}, Unreal 4K~\cite{tosi2021smd}, Virtual KITTI~\cite{cabon2020virtual}, Waymo~\cite{sun2020scalability}, WildRGBD~\cite{xia2024rgbd}.
These datasets cover both synthetic and real-world content, indoor and outdoor environments, as well as static and dynamic scenes. Such a diverse composition ensures strong generalization capability for \ours.
See appendix for the full dataset details.

\noindent \textbf{Implementation Details.}
Our model architecture follows VGGT~\cite{wang2025vggt} with $L = 24$ AA blocks.
Camera GeoAdapter has ($L+1$) independent camera encoders, each consisting of a single linear layer.
The depth GeoAdapter has only one depth encoder, which consists of a single convolutional layer with a kernel size of 14, patchifying the depth maps into the same dimension as the spatial tokens.
Our lightweight GeoAdapter introduces only 26.8M additional parameters.
The training runs end-to-end on 32 NVIDIA A100 GPUs over ten days.
Gradient checkpointing is used to optimize memory usage.

\noindent \textbf{Metrics.} For \textit{depth evaluation}, we use absolute relative error (Abs Rel) and percentage of inlier points $\delta < 1.25$ following~\cite{hu2025depthcrafter}.
For \textit{camera pose estimation}, we use relative rotation accuracy (RRA) and relative translation accuracy (RTA) as the metrics following~\cite{wang2025vggt}, which calculate the relative angular errors in rotation and translation, respectively, for each image pair. 
In addition, AUC is the area under the accuracy threshold curve of the minimum values between RRA and RTA across varying thresholds.
For \textit{3D reconstruction}, we use the standard metrics of Accuracy (Acc), Completeness (Comp), and Normal Consistency (NC), following~\cite{wang20243d,wang2025cut3r,wang2024dust3r}.

\subsection{Auxiliary Information Guidance}

We perform a zero-shot evaluation on the Sintel dataset~\cite{butler2012naturalistic} to showcase the effect of injecting auxiliary information at arbitrary ratios on camera pose and depth prediction.
Specifically, we randomly sample 10 images from 100 scenes. All images are resized to a fixed width of 518 pixels, with the aspect ratio adjusted to match the closest ratio used during training. The reported results are averaged over all samples.
As shown in Table~\ref{tab:maintable}, our method outperforms the VGGT baseline even in the auxiliary-free setting, achieving a 0.65 reduction in \(\delta<1.25\) for depth estimation and a 0.46 gain in RRA@5$^\circ$ for pose estimation. 
Moreover, our method demonstrates strong scalability, showing consistent improvements across all metrics as the proportion of available ground-truth camera or depth information increases. 
For instance, incorporating only 30\% of the depth information reduces the Abs.~Rel error by 69.71\%.
Furthermore, when incorporating auxiliary depth information, our method also achieves notable gains in pose estimation. 
For example, with 100\% of the depth information injected, the pose estimation improves by 6.33 in AUC@30$^\circ$. 
This further demonstrates that our approach enhances the accuracy of spatial representations through auxiliary cues, rather than merely learning a direct mapping from input to output.
Qualitative results in Fig.~\ref{fig:ablation} further confirm this observation: providing auxiliary pose information not only leads to more accurate pose predictions but also yields more faithful geometric representations. 
Meanwhile, supplying auxiliary depth information enhances geometric details and improves inter-view pose alignment accuracy.

\subsection{Depth Estimation}
\label{sec:depth}
\noindent \textbf{Mono-Depth Estimation.} We first evaluate monocular depth estimation on Sintel~\cite{butler2012naturalistic}, Bonn~\cite{palazzolo2019refusion}, and NYU-v2~\cite{silberman2012indoor} datasets following the previous benchmark~\cite{zhang2024monst3r,wang2025continuous}, which cover dynamic and static, indoor and outdoor, realistic and synthetic data. 
These datasets are not used for training and are suitable for benchmarking the zero-shot performance across different domains.

As shown in Table~\ref{tab:single_frame_depth}, our method achieves superior performance on both the Sintel and NYU-v2 datasets, surpassing the baseline by 0.5 and 1.0 in $\delta\!<\!1.25$ metric even without any auxiliary information.
Moreover, compared to Pow3R~\cite{jang2025pow3r}, our method achieves greater performance gains when incorporating auxiliary depth information (\textit{w/} D), reaching nearly perfect accuracy (99.9\% in $\delta\!<\!1.25$) on both the Bonn and NYU-v2 datasets. 
This demonstrates that our GeoAdapter effectively injects auxiliary information into the spatial representations, thereby enhancing the performance of downstream tasks.

\begin{table}[t]
\centering
\caption{\textbf{Single-frame Depth Evaluation.} We report the performance on Sintel, Bonn, and NYU-v2 (static) datasets. 
`\textit{w/} D' indicates that 100\% depth maps are provided.
}
\vspace{-1mm}
\resizebox{\linewidth}{!}{
\begin{tabular}{@{}>{\centering\arraybackslash}l|
                >{\centering\arraybackslash}c >{\centering\arraybackslash}c|
                >{\centering\arraybackslash}c >{\centering\arraybackslash}c|
                >{\centering\arraybackslash}c >{\centering\arraybackslash}c@{}}
\toprule
\multicolumn{1}{c|}{\multirow{3}{*}{\textbf{Method}}}
  & \multicolumn{2}{c}{\textbf{Sintel}} 
  & \multicolumn{2}{c}{\textbf{Bonn}} 
  & \multicolumn{2}{c}{\textbf{NYU-v2}} \\
\cmidrule(lr){2-3} \cmidrule(lr){4-5} \cmidrule(lr){6-7}
  & {\footnotesize Abs Rel $\downarrow$} & {\footnotesize $\delta$\textless{}$1.25\uparrow$} 
  & {\footnotesize Abs Rel $\downarrow$} & {\footnotesize $\delta$\textless{}$1.25\uparrow$} 
  & {\footnotesize Abs Rel $\downarrow$} & {\footnotesize $\delta$\textless{}$1.25\uparrow$} \\
\midrule
VGGT~\cite{wang2025vggt} & 0.271 & \multicolumn{1}{c|}{67.7} & 0.053 & \multicolumn{1}{c|}{97.3} & 0.060 & 94.8 \\
Fast3R~\cite{yang2025fast3r} & 0.502 & \multicolumn{1}{c|}{52.8} & 0.192 & \multicolumn{1}{c|}{77.3} & 0.099 & 88.9 \\
DUSt3R~\cite{wang2024dust3r} & 0.424 & \multicolumn{1}{c|}{58.7} & 0.141 & \multicolumn{1}{c|}{82.5} & 0.080 & 90.7 \\
MASt3R~\cite{leroy2024mast3r} & 0.340 & \multicolumn{1}{c|}{60.4} & 0.142 & \multicolumn{1}{c|}{82.0} & 0.129 & 84.9 \\
MonST3R~\cite{zhang2024monst3r} & 0.358 & \multicolumn{1}{c|}{54.8} & 0.076 & \multicolumn{1}{c|}{93.9} & 0.102 & 88.0 \\
Spann3R~\cite{wang20243d} & 0.470 & \multicolumn{1}{c|}{53.9} & 0.118 & \multicolumn{1}{c|}{85.9} & 0.122 & 84.9 \\
CUT3R~\cite{wang2025cut3r} & 0.428 & \multicolumn{1}{c|}{55.4} & 0.063 & \multicolumn{1}{c|}{96.2} & 0.086 & 90.9 \\ \midrule
Pow3R~\cite{jang2025pow3r} & 0.464 & \multicolumn{1}{c|}{54.8} & 0.132 & \multicolumn{1}{c|}{84.2} & 0.094 & 89.6 \\
Pow3R w/ D~\cite{jang2025pow3r} & \underline{0.150} & \multicolumn{1}{c|}{\underline{87.4}} & \underline{0.009} & \multicolumn{1}{c|}{\underline{99.7}} & \underline{0.009} & \underline{99.8} \\
\textbf{\ours} & 0.250 & \multicolumn{1}{c|}{68.2} & 0.064 & \multicolumn{1}{c|}{95.5} & 0.058 & 95.8 \\
\textbf{\ours w/ D} & \bf0.107 & \multicolumn{1}{c|}{\bf90.2} & \bf0.008 & \multicolumn{1}{c|}{\bf99.9} & \bf0.008 & \bf99.9 \\ \bottomrule
\end{tabular}
}
\label{tab:single_frame_depth}
\end{table}

\noindent \textbf{Multi-View Depth Estimation.}
Following RobustMVD~\cite{schroppel2022benchmark}, we evaluate the multi-view depth performance on the ScanNet~\cite{dai2017scannet}, ETH3D~\cite{schops2017multi}, DTU~\cite{aanaes2016large}, and Tanks and Temples~\cite{knapitsch2017tanks} datasets. 

The detailed results are presented in Table~\ref{tab:multiviewDepth}, where our method consistently outperforms prior approaches in both accuracy and robustness. 
Without any auxiliary information, it already achieves superior results on ScanNet and comparable or better scores on ETH3D and DTU. 
Furthermore, incorporating auxiliary depth supervision (\textit{w/} D) yields a pronounced improvement across all datasets, reaching near-perfect correlation on ETH3D and DTU.

\begin{table*}[t]
\caption{\textbf{Multi-view Depth Evaluation.}  (Parentheses) denote training on data from the same domain. “K”, “RT”, and “D” denote intrinsic, relative pose, and depth information, respectively. 
The best and second best results are \textbf{bold} and \underline{underlined} respectively.}
\label{tab:multiviewDepth}
\vspace{-1mm}
\centering
\footnotesize
\resizebox{0.9\linewidth}{!}{
\begin{tabular}{@{}l|cc|cccccccccc@{}}
\toprule
\multicolumn{1}{c|}{\multirow{2}{*}{\textbf{Method}}} & \multicolumn{1}{c|}{\textbf{GT}} & \multicolumn{1}{c|}{\textbf{Align}} & \multicolumn{2}{c}{\textbf{ScanNet~\cite{dai2017scannet}}} & \multicolumn{2}{c}{\textbf{ETH3D~\cite{schops2017multi}}} & \multicolumn{2}{c}{\textbf{DTU~\cite{aanaes2016large}}} & \multicolumn{2}{c}{\textbf{T\&T~\cite{knapitsch2017tanks}}} & \multicolumn{2}{c}{\textbf{Average}} \\ \cmidrule(l){4-13} 
 & \multicolumn{1}{c|}{\textbf{Range}} & \multicolumn{1}{c|}{\textbf{Method}} & \textbf{rel$\downarrow$} & \textbf{$\tau\uparrow$} & \textbf{rel$\downarrow$} & \textbf{$\tau\uparrow$} & \textbf{rel$\downarrow$} & \textbf{$\tau\uparrow$} & \textbf{rel$\downarrow$} & \multicolumn{1}{c|}{\textbf{$\tau\uparrow$}} & \textbf{rel$\downarrow$} & \textbf{$\tau\uparrow$} \\ \midrule
COLMAP~\cite{schonberger2016pixelwise, schonberger2016structure} (K+RT) & \multicolumn{1}{c|}{$\times$} & $\times$ & 14.6 & 34.2 & 16.4 & 55.1 & 0.7 & 96.5 & 2.7 & \multicolumn{1}{c|}{95.0} & 8.6 & 70.2 \\
COLMAP Dense~\cite{schonberger2016pixelwise, schonberger2016structure} (K+RT) & \multicolumn{1}{c|}{$\times$} & $\times$ & 38.0 & 22.5 & 89.8 & 23.2 & 20.8 & 69.3 & 25.7 & \multicolumn{1}{c|}{76.4} & 43.6 & 47.9 \\
MVSNet~\cite{yao2018mvsnet} (K+RT) & \multicolumn{1}{c|}{$\checkmark$} & $\times$ & 22.7 & 20.9 & 21.6 & 35.6 & (1.8) & (86.7) & 6.5 & \multicolumn{1}{c|}{74.6} & 12.3 & 11.1 \\
Vis-MVSNet~\cite{zhang2020visibility} (K+RT) & \multicolumn{1}{c|}{$\checkmark$} & $\times$ & 8.9 & 33.5 & 10.8 & 43.3 & 1.8 & 87.4 & 4.1 & \multicolumn{1}{c|}{7.2} & 6.4 & 42.9 \\
MVS-Former++ (K+RT) & \multicolumn{1}{c|}{$\checkmark$} & $\times$ & 15.2 & 21.9 & 21.4 & 32.5 & (1.2) & (91.9) & 7.6 & \multicolumn{1}{c|}{71.5} & 10.8 & 8.5 \\
CER-MVS~\cite{cao2024mvsformer++} (K+RT) & \multicolumn{1}{c|}{$\times$} & $\times$ & 21.1 & 24.3 & 11.7 & 47.5 & 4.1 & 71.3 & 6.4 & \multicolumn{1}{c|}{82.1} & 10.8 & 56.3 \\ \midrule
DUSt3R~\cite{wang2024dust3r} & \multicolumn{1}{c|}{$\times$} & med & (3.1) & (71.8) & 3.0 & 76.0 & 3.9 & 68.6 & 3.3 & \multicolumn{1}{c|}{75.1} & 3.3 & 72.9 \\
Pow3R~\cite{jang2025pow3r} & \multicolumn{1}{c|}{$\times$} & med & (3.2) & (68.8) & 3.0 & 74.7 & 3.0 & 74.3 & 3.3 & \multicolumn{1}{c|}{76.6} & 3.1 & 73.6 \\
VGGT~\cite{wang2025vggt} & \multicolumn{1}{c|}{$\times$} & med & (3.7) & (70.0) & 1.7 & 87.2 & 0.9 & 95.4 & 1.7 & \multicolumn{1}{c|}{90.6} & 2.0 & 85.8 \\
\textbf{\ours} & \multicolumn{1}{c|}{$\times$} & med  & (3.6) & (72.3) & 1.8 & 87.5 & 1.1 & 93.9 & 1.8 & \multicolumn{1}{c|}{90.0} & 2.1 & 85.9 \\ \midrule
Pow3R~\cite{jang2025pow3r} w/ (K+RT) & \multicolumn{1}{c|}{$\times$} & med & (3.1) & (71.4) & 2.8 & 77.1 & 1.5 & 91.1 & 3.2 & \multicolumn{1}{c|}{78.2} & 2.7 & 79.5 \\
\textbf{\ours w/ (K+RT)} & \multicolumn{1}{c|}{$\times$} & med & (3.7) & (72.2) & 1.8 & 87.8 & 1.2 & 93.6 &  1.8 &  \multicolumn{1}{c|}{89.9} & 2.1 & 85.9 \\
\textbf{\ours w/ D} & \multicolumn{1}{c|}{$\times$} & med & (\underline{2.3}) & (\underline{85.6}) & \underline{0.5} & \underline{98.7}  & \underline{0.3}  & \textbf{99.5} & \underline{0.9} & \multicolumn{1}{c|}{\underline{95.5}} & \underline{1.0} & \underline{94.8} \\
\textbf{\ours w/ (K+RT+D)} & \multicolumn{1}{c|}{$\times$} & med & (\textbf{2.2}) & (\textbf{86.7}) & \textbf{0.5} & \textbf{98.7} & \textbf{0.3} & \underline{99.4} & \textbf{0.9} & \multicolumn{1}{c|}{\textbf{95.6}} & \textbf{1.0} & \textbf{95.1} \\ \bottomrule
\end{tabular}}
\end{table*}

\begin{figure*}[!t]
\begin{center}
\includegraphics[width=1\linewidth]{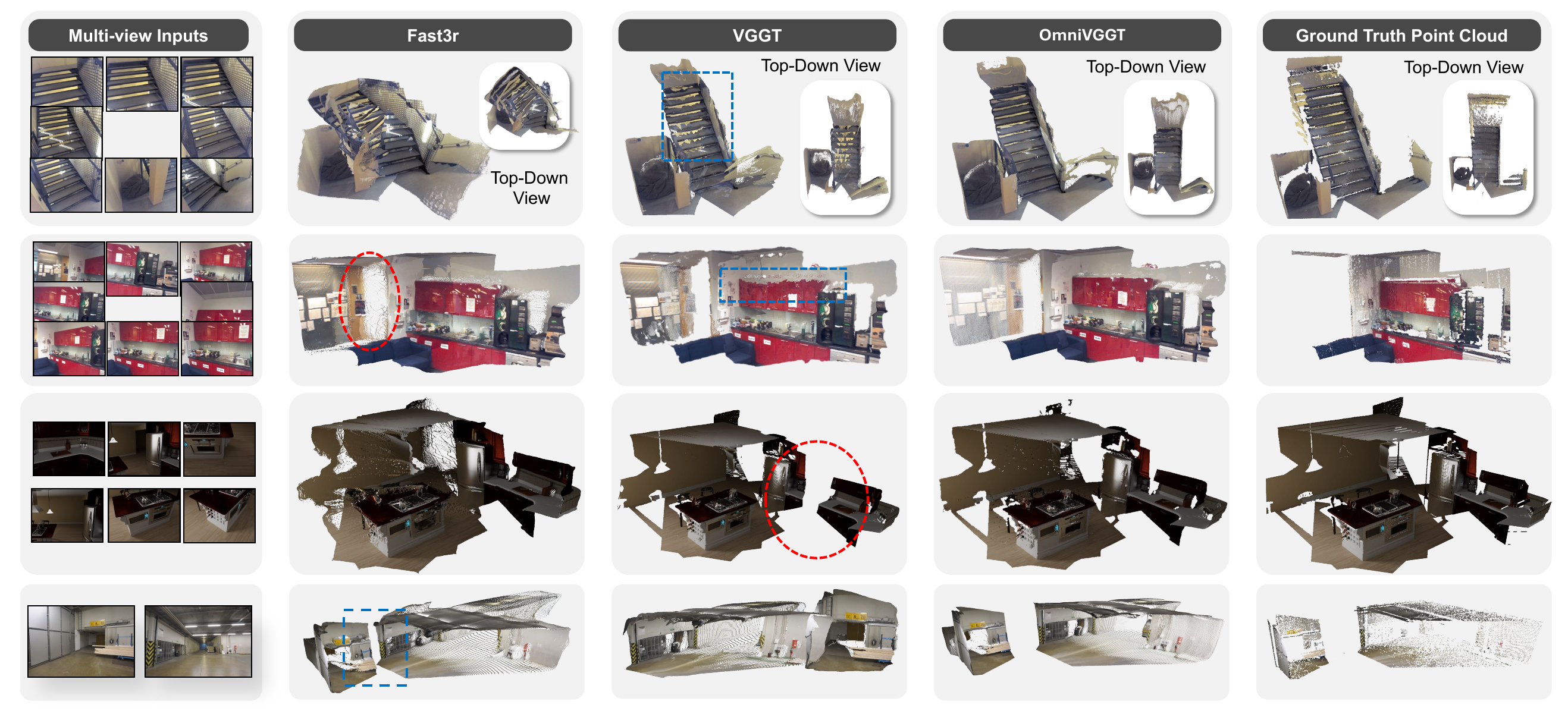}
\caption{\textbf{Visual Comparisons on 7-Scenes~\cite{shotton2013scene}, NRGBD~\cite{azinovic2022neural}, and ETH3D~\cite{schops2017multi} datasets.} \ours exhibits accurate spatial relationships and geometric consistency, even in extremely challenging cases. 
More examples can be found in the appendix.}
\label{fig:visualization1}
\end{center}
\vspace{-7mm}
\end{figure*}

\subsection{Camera Pose Estimation}
We evaluate \ours for camera pose prediction on Co3Dv2~\cite{reizenstein2021common} and RealEstate10K~\cite{zhou2018stereo}
Specifically, we randomly select 10 images from each scene and evaluate them using the standard AUC@$30^{\circ}$ metric following~\cite{wang2023posediffusion}.

As shown in Table~\ref{tab:camerapose}, in the RGB-only setting, \ours outperforms the state-of-the-art method VGGT, and significantly surpasses all other baselines. Furthermore, when auxiliary inputs are available, \ours demonstrates consistent performance gains, outperforming the advanced auxiliary-aware method Pow3R, by up to 16\% on Re10K.
Remarkably, this superior performance is achieved with unprecedented flexibility and efficiency: \ours is the first to accept an arbitrary number of auxiliary inputs, yet it operates approximately 30$\times$ faster than Pow3R.

\begin{table}[t]
\caption{\textbf{Camera Pose Estimation on the RealEstate10K~\cite{zhou2018stereo} and CO3Dv2~\cite{reizenstein2021common} datasets.} Pro means using Procrustes alignment. `()' means not trained on CO3D. }
\label{tab:camerapose}
\vspace{-1mm}
\footnotesize
\centering
\resizebox{\linewidth}{!}{
\begin{tabular}{@{}l|c|c|c@{}}
\toprule
\multicolumn{1}{c|}{\multirow{2}{*}{\textbf{Method}}} & \textbf{Re10K \textit{unseen}} & \textbf{CO3Dv2} & \multirow{2}{*}{\textbf{Time}} \\
 & \textbf{AUC@$30^{\circ}(\%) \uparrow$} & \textbf{AUC@$30^{\circ}(\%) \uparrow$} &  \\ \midrule
COLMAP+SPSG~\cite{sarlin2020superglue} & 45.2 & 25.3 & $\sim$ 15s \\
PixSfM~\cite{lindenberger2021pixel} & 49.4 & 30.1 & $>$ 20s \\
PoseDiff~\cite{wang2023posediffusion} & 48.0 & 66.5 & $\sim$ 7s \\
DUSt3R~\cite{wang2024dust3r} & 67.7 & 76.7 & $\sim$ 7s \\
MASt3R~\cite{leroy2024mast3r} & 76.4 & 81.8 & $\sim$ 9s \\
VGGSfM v2~\cite{wang2024vggsfm} & 78.9 & 83.4 & $\sim$ 10s \\ \midrule
MV-DUSt3R~\cite{tang2025mv} & 71.3 & (69.5) & $\sim$ 0.6s \\
CUT3R~\cite{wang2025cut3r} & 75.3 & 82.8 & $\sim$ 0.6s \\
FLARE~\cite{zhang2025flare} & 78.8 & 83.3 & $\sim$ 0.5s \\
Fast3R~\cite{yang2025fast3r} & 72.7 & 82.5 & $\sim$ 0.2s \\
VGGT~\cite{wang2025vggt} & 85.3 & 88.2 & $\sim$ 0.2s \\
Pow3R (PnP)~\cite{hartley2003multiple} & 62.5 & 78.5 & $>$ 7s \\
Pow3R (Pro)~\cite{jang2025pow3r} & 62.5 & 78.5 & $>$ 7s \\
\textbf{\ours} & 85.9 & 88.4 & \textbf{$\sim$ 0.2s} \\ \midrule
Pow3R w/ K (Pro)~\cite{jang2025pow3r} & 72.5 & 82.2 & $>$ 7s \\
\textbf{\ours w/ D} & \underline{85.5} & \underline{91.3} & $\sim$ 0.2s \\ 
\textbf{\ours w/ K+RT} & \textbf{88.5} & \textbf{93.4} & \textbf{$\sim$ 0.2s} \\ \bottomrule
\end{tabular}}
\end{table}

\subsection{3D Reconstrunction}
We evaluate scene-level 3D reconstruction on the 7-Scenes~\cite{shotton2013scene} benchmark following previous works~\cite{wang20243d,wang2025cut3r,wang2024dust3r}.
In this benchmark, each scene consists of 3–5 sparsely captured frames with minimal or no overlap between views. 

As shown in Table~\ref{tab:7scenereconstruct}, \ours achieves competitive performance with existing state-of-the-art VGGT in the RGB-only setting. 
Furthermore, it significantly boosts performance when auxiliary inputs (e.g., depth and camera parameters) are available, substantially outperforming all existing baselines.
Notably, the inclusion of camera parameters (\ours \textit{w/} (K+RT)) yields a remarkable 65.4\% performance gain (metric improving from 0.104 to 0.036) over the baseline \ours. 
We attribute this to the extreme image sparsity of the 7-Scenes dataset, as illustrated in Fig.~\ref{fig:visualization1}. 
This sparsity makes from-scratch camera pose estimation exceptionally difficult and creates a severe performance bottleneck. 
Our method effectively overcomes this bottleneck by leveraging the accessible camera pose information, demonstrating its robustness in handling sparse and non-overlapping viewpoints.

\begin{table}[t]
\caption{\textbf{3D reconstruction on the 7-scenes~\cite{shotton2013scene} datasets.} "K", "RT", and "D" denote intrinsic, relative pose, and depth information, respectively. }
\label{tab:7scenereconstruct}
\vspace{-1mm}
\centering
\footnotesize
\resizebox{\linewidth}{!}{
\begin{tabular}{@{}l|cc|cc|cc@{}}
\toprule
\multicolumn{1}{c|}{\multirow{2}{*}{\textbf{Method}}} & \multicolumn{2}{c|}{\textbf{Acc$\downarrow$}} & \multicolumn{2}{c|}{\textbf{Comp$\downarrow$}} & \multicolumn{2}{c|}{\textbf{NC$\uparrow$}}  \\ \cmidrule(lr){2-7}
 & \textbf{Mean} & \textbf{Med.} & \textbf{Mean} & \textbf{Med.} & \textbf{Mean} & \textbf{Med.}   \\ \midrule
VGGT~\cite{wang2025vggt} & \textbf{0.087} & \underline{0.039} & \textbf{0.091} & \underline{0.039} & \textbf{0.787} & \textbf{0.890}  \\
Fast3R~\cite{yang2025fast3r} & 0.164 & 0.108 & 0.163 & 0.080 & 0.686 & 0.775  \\
DUSt3R-GA~\cite{wang2024dust3r} & 0.146 & 0.077 & 0.181 & 0.067 & 0.736 & 0.839  \\
MASt3R-GA~\cite{leroy2024mast3r} & 0.185 & 0.081 & 0.180 & 0.069 & 0.701 & 0.792  \\
MonST3R-GA~\cite{zhang2024monst3r} & 0.248 & 0.185 & 0.266 & 0.167 & 0.672 & 0.759  \\
Spann3R~\cite{wang20243d} & 0.298 & 0.226 & 0.205 & 0.112 & 0.650 & 0.730  \\
SLAM3R~\cite{liu2025slam3r} & 0.287 & 0.155 & 0.226 & 0.066 & 0.644 & 0.720  \\
CUT3R~\cite{wang2025cut3r} & 0.126 & 0.047 & 0.154 & 0.031 & 0.727 & 0.834  \\
\textbf{\ours} & \underline{0.104} & \textbf{0.037} & \underline{0.112} & \textbf{0.031} & \underline{0.763} & \underline{0.875}  \\ \midrule[0.8pt]
\textbf{\ours  w/ D }  & 0.085 & 0.034 & 0.085 & 0.027 & \underline{0.789} & \underline{0.894}  \\ 
\textbf{\ours w/ (K+RT)} & \underline{0.037} & \underline{0.017} & \underline{0.049} & \underline{0.019} &  0.778 & 0.893  \\ 
\textbf{\ours w/ (K+RT+D)} & \textbf{0.036} & \textbf{0.017} & \textbf{0.036} & \textbf{0.017} &  \textbf{0.810} & \textbf{0.912}  \\
\bottomrule
\end{tabular}}
\end{table}

\subsection{Application on Vision-Language-Action Model}
\ours generates richer and more representative 3D spatial tokens by progressively incorporating auxiliary inputs like depth and pose. This inherent spatial awareness is vital for the VLA model~\cite{peng2023kosmos,li2023generalist}, which is required to interact with the physical 3D world and, critically, predict absolute poses for robotic manipulation. 
Specifically, as shown in Fig. 1, we build upon the Kosmos-VLA 1.6B model~\cite{peng2023kosmos} with an FCN action head, where our spatial tokens are injected and fused with the VLM tokens to enhance spatial reasoning. 
We then finetune the VLA model and evaluate it on the CALVIN dataset~\cite{mees2022calvin}. 
As shown in Table~\ref{tab:abl_rgbd}, under the ABCD$\rightarrow$D setting, our method with auxiliary depth inputs outperforms approaches that incorporate 3D priors via a point encoder by 0.04 in the Avg. Len metric. Under the zero-shot ABC$\rightarrow$D setting, our RGB-only variant also surpasses the Kosmos $w/$ RGB baseline by a large margin of 0.43 in Avg. Len. These results demonstrate that our approach learns richer spatial representations and can be further enhanced by auxiliary inputs to improve robotic manipulation accuracy. See supplement for rollout examples.

\begin{table}
\caption{Performance comparison of different modality input on CALVIN~\cite{mees2022calvin} benchmark. Kosmos-VLA (\textit{w/ rgb-d}) is a point cloud-based version with a lightweight point cloud encoder~\citep{ze2025manipulation3d} while retaining other parts.}
\vspace{-1mm}
\resizebox{\linewidth}{!}{
        \begin{tabular}{l l c c c c c c}
            \toprule
            \multicolumn{1}{c}{\textbf{Method}}  & \textbf{Task} & \multicolumn{5}{c}{\textbf{Tasks Completed in a Row (\%)}} & \textbf{Avg. Len. $\uparrow$} \\
            \cmidrule(lr){3-7}
            &  & 1 & 2 & 3 & 4 & 5 & \\
            \midrule
    
            Kosmos-VLA (w/ rgb)& ABCD$\rightarrow$D  & 92.9        & 85.4        & 79.4            & 74.4          & 68.1         & 4.00          \\
            Kosmos-VLA (w/ rgb-d)  & ABCD$\rightarrow$D  & 93.4        & 85.8        & 80.5            & 75.3          & 69.2          & 4.04          \\
            \textbf{Ours (w/ rgb})  & ABCD$\rightarrow$D & \textbf{93.8} & 86.6        & 81.0          & 75.5          & 69.5 & 4.07          \\
            \textbf{Ours (w/ rgb-d)} & ABCD$\rightarrow$D & 93.7 & \textbf{86.8} & \textbf{81.4} & \textbf{76.7} & \textbf{70.2}          & \textbf{4.08} \\
            
            \midrule
            Kosmos-VLA (w/ rgb)& ABC$\rightarrow$D  & 90.1        & 79.1        & 69.2            & 59.6          & 50.9          & 3.49          \\
            
            Kosmos-VLA (w/ rgb-d) & ABC$\rightarrow$D  & 93.6    & 86.0     & 78.6          & \textbf{72.9} & \textbf{64.8} & \textbf{3.97}          \\
            \textbf{Ours (w/ rgb}) & ABC$\rightarrow$D  & 93.8          & 86.9 & 77.9          & 70.3          & 62.2          & 3.92          \\
            \textbf{Ours (w/ rgb-d}) & ABC$\rightarrow$D & \textbf{95.1} & \textbf{87.7}          & \textbf{79.2} & 70.8          & 63.0          & 3.96  \\
            
            \bottomrule
        \end{tabular}
        }
        \label{tab:abl_rgbd}
        \vspace{-0.1cm}
\end{table}

\subsection{Architecture Ablation}
\begin{figure}[!t]
\begin{center}
\includegraphics[width=1\linewidth]{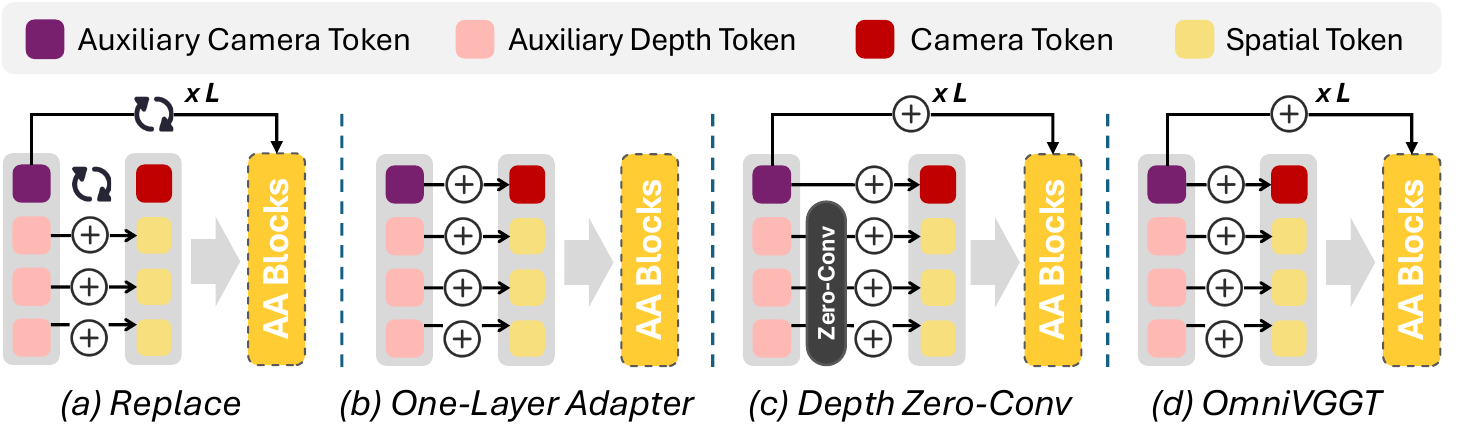}
\vspace{-6mm}
\caption{\textbf{Visualization of our GeoAdapter module in ablation.}  }
\label{fig:abl_vis}
\end{center}
\vspace{-6mm}
\end{figure}

We conduct a comprehensive ablation study on GeoAdapter to thoroughly analyze its working mechanism.
As shown in Fig.~\ref{fig:abl_vis}, we evaluate three variants: 
(a) \textit{Replace}, where the original camera tokens are directly replaced by auxiliary camera tokens in each AA block. 
(b) \textit{One-Layer Adapter}, where auxiliary camera tokens are only injected once before the AA blocks. 
(c) \textit{Depth ZeroConv}, where a ZeroConv layer is used for depth information injection.

\begin{figure}[!t]
\begin{center}
\includegraphics[width=1\linewidth]{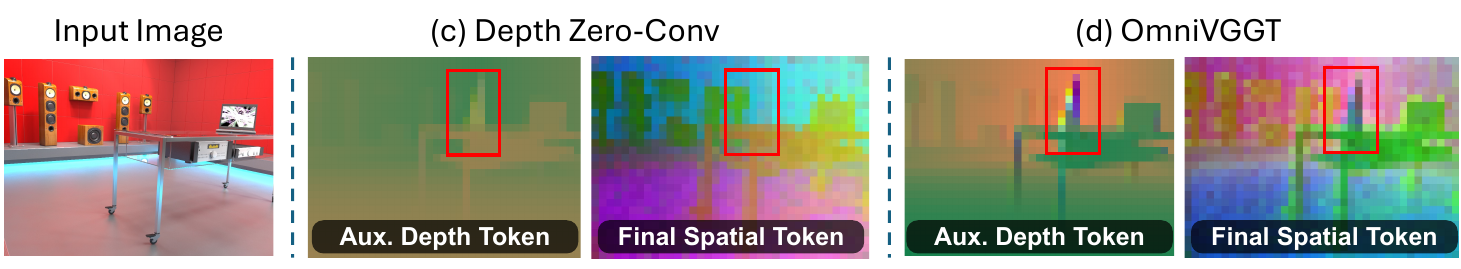}
\vspace{-6mm}
\caption{\textbf{Token visualization of our GeoAdapter module. }Discriminative areas are highlighted with '{\color[HTML]{CB0000}$\Box$}'.}
\label{fig:abl_pca_vis}
\end{center}
\vspace{-8mm}
\end{figure}

As shown in Table~\ref{tab:ablation}, variant \textit{(a)} yields suboptimal results in both the RGB-only setting and the full auxiliary information setting, while variant \textit{(b)} is insufficient to strengthen the correct prior knowledge in the full auxiliary information setting.
Regarding \textit{(c)}, the model tends to interpret the auxiliary depth information as noise, leading to limited performance improvement.
As illustrated by the PCA visualization in Fig.~\ref{fig:abl_pca_vis}, the \textit{(d)} configuration preserves more complete and discriminative auxiliary depth information than \textit{(c)}, leading to a more accurate spatial representation.

\begin{table}[t]
\caption{\textbf{Ablation of GeoAdapter architectures.} We compare different GeoAdapter designs on the Sintel~\cite{butler2012naturalistic} dataset. 
}\label{tab:ablation}
\vspace{-1mm}
\centering
\footnotesize
\resizebox{\linewidth}{!}{
\begin{tabular}{@{}l|cc|ccc@{}}
\toprule
\multicolumn{1}{c|}{\multirow{2}{*}{\textbf{Architecture}}} & \multicolumn{2}{c|}{\textbf{Depth}} & \multicolumn{3}{c}{\textbf{Camera}} \\ \cmidrule(l){2-6} 
 & \textbf{Abs Rel$\downarrow$} & \textbf{$\delta<1.25\uparrow$} & RRA@$5^{\circ}\uparrow$ & RTA@$5^{\circ}\uparrow$ & AUC@$30^{\circ}\uparrow$ \\ \midrule
\multicolumn{6}{l}{\textit{w/o (K+RT+D) Auxiliary Information}} \\ \cmidrule(l){1-6}

\textbf{(a) Replace} & 0.845 & 64.74 & 93.40 & 30.88 & 64.74 \\

\textbf{(b) One-Layer Adpter} & 0.604 & 68.74 & \textbf{96.78} & 44.92 & 68.74 \\

\textbf{(c) Depth ZeroConv} & 0.569 & 70.71 & 96.44 & 51.86 & 69.70 \\

\textbf{(d) \ours} & \textbf{0.558} & \textbf{71.46} & 96.15 & \textbf{54.01} & \textbf{70.83} \\ \midrule

\multicolumn{6}{l}{\textit{w/ (K+RT+D) Auxiliary Information}} \\ \cmidrule(l){1-6}

\textbf{(a) Replace} & 0.655 & 82.96 & 97.08 & 57.61 & 77.83 \\

\textbf{(b) One-Layer Adpter} & 0.133 & 85.65 & 99.97 & 60.89 & 81.66 \\

\textbf{(c) Depth ZeroConv} & 0.505 & 71.11 & 99.72 & 71.66 & 84.12 \\

\textbf{(d) \ours} & \textbf{0.106} & \textbf{85.95} & \textbf{99.97} & \textbf{76.33} & \textbf{85.99} \\ \bottomrule

\end{tabular}}

\end{table}

\section{Conclusion}

We propose \ours, a unified feed-forward model capable of handling diverse input modalities for various 3D tasks. 
Our model flexibly accommodates casual inputs, including arbitrary numbers of images, auxiliary information, and input combinations. 
It achieves state-of-the-art performance across multiple tasks (from reconstruction to robotics) while maintaining high inference efficiency.
Our findings demonstrate that integrating multi-modalities information within a unified architecture can significantly advance the development of 3D foundation models.
\newpage
{
    \small
    \bibliographystyle{ieeenat_fullname}
    \bibliography{main}

@String(CVPR= {IEEE Conf. Comput. Vis. Pattern Recog.})

@String(ECCV= {Eur. Conf. Comput. Vis.})

@String(TOG= {ACM Trans. Graph.})

@String(CVPR  = {CVPR})

@String(ECCV  = {ECCV})

@String(TOG   = {ACM TOG})

@inproceedings{wang2024vggsfm,
  title={Vggsfm: Visual geometry grounded deep structure from motion},
  author={Wang, Jianyuan and Karaev, Nikita and Rupprecht, Christian and Novotny, David},
  booktitle={Proceedings of the IEEE/CVF conference on computer vision and pattern recognition},
  pages={21686--21697},
  year={2024}
}

@inproceedings{wang2025vggt,
  title={Vggt: Visual geometry grounded transformer},
  author={Wang, Jianyuan and Chen, Minghao and Karaev, Nikita and Vedaldi, Andrea and Rupprecht, Christian and Novotny, David},
  booktitle={Proceedings of the Computer Vision and Pattern Recognition Conference},
  pages={5294--5306},
  year={2025}
}

@article{oquab2023dinov2,
  title={Dinov2: Learning robust visual features without supervision},
  author={Oquab, Maxime and Darcet, Timoth{\'e}e and Moutakanni, Th{\'e}o and Vo, Huy and Szafraniec, Marc and Khalidov, Vasil and Fernandez, Pierre and Haziza, Daniel and Massa, Francisco and El-Nouby, Alaaeldin and others},
  journal={arXiv preprint arXiv:2304.07193},
  year={2023}
}

@inproceedings{ranftl2021vision,
  title={Vision transformers for dense prediction},
  author={Ranftl, Ren{\'e} and Bochkovskiy, Alexey and Koltun, Vladlen},
  booktitle={Proceedings of the IEEE/CVF international conference on computer vision},
  pages={12179--12188},
  year={2021}
}

@article{baruch2021arkitscenes,
  title={Arkitscenes: A diverse real-world dataset for 3d indoor scene understanding using mobile rgb-d data},
  author={Baruch, Gilad and Chen, Zhuoyuan and Dehghan, Afshin and Dimry, Tal and Feigin, Yuri and Fu, Peter and Gebauer, Thomas and Joffe, Brandon and Kurz, Daniel and Schwartz, Arik and others},
  journal={arXiv preprint arXiv:2111.08897},
  year={2021}
}

@inproceedings{yao2020blendedmvs,
  title={Blendedmvs: A large-scale dataset for generalized multi-view stereo networks},
  author={Yao, Yao and Luo, Zixin and Li, Shiwei and Zhang, Jingyang and Ren, Yufan and Zhou, Lei and Fang, Tian and Quan, Long},
  booktitle={Proceedings of the IEEE/CVF conference on computer vision and pattern recognition},
  pages={1790--1799},
  year={2020}
}

@inproceedings{ling2024dl3dv,
  title={Dl3dv-10k: A large-scale scene dataset for deep learning-based 3d vision},
  author={Ling, Lu and Sheng, Yichen and Tu, Zhi and Zhao, Wentian and Xin, Cheng and Wan, Kun and Yu, Lantao and Guo, Qianyu and Yu, Zixun and Lu, Yawen and others},
  booktitle={Proceedings of the IEEE/CVF Conference on Computer Vision and Pattern Recognition},
  pages={22160--22169},
  year={2024}
}

@inproceedings{karaev2023dynamicstereo,
  title={Dynamicstereo: Consistent dynamic depth from stereo videos},
  author={Karaev, Nikita and Rocco, Ignacio and Graham, Benjamin and Neverova, Natalia and Vedaldi, Andrea and Rupprecht, Christian},
  booktitle={Proceedings of the IEEE/CVF Conference on Computer Vision and Pattern Recognition},
  pages={13229--13239},
  year={2023}
}

@inproceedings{roberts2021hypersim,
  title={Hypersim: A photorealistic synthetic dataset for holistic indoor scene understanding},
  author={Roberts, Mike and Ramapuram, Jason and Ranjan, Anurag and Kumar, Atulit and Bautista, Miguel Angel and Paczan, Nathan and Webb, Russ and Susskind, Joshua M},
  booktitle={Proceedings of the IEEE/CVF international conference on computer vision},
  pages={10912--10922},
  year={2021}
}

@inproceedings{greff2022kubric,
  title={Kubric: A scalable dataset generator},
  author={Greff, Klaus and Belletti, Francois and Beyer, Lucas and Doersch, Carl and Du, Yilun and Duckworth, Daniel and Fleet, David J and Gnanapragasam, Dan and Golemo, Florian and Herrmann, Charles and others},
  booktitle={Proceedings of the IEEE/CVF conference on computer vision and pattern recognition},
  pages={3749--3761},
  year={2022}
}

@inproceedings{arnold2022map,
  title={Map-free visual relocalization: Metric pose relative to a single image},
  author={Arnold, Eduardo and Wynn, Jamie and Vicente, Sara and Garcia-Hernando, Guillermo and Monszpart, Aron and Prisacariu, Victor and Turmukhambetov, Daniyar and Brachmann, Eric},
  booktitle={European Conference on Computer Vision},
  pages={690--708},
  year={2022},
  organization={Springer}
}

@inproceedings{ramakrishnan2021hm3d,
  title={Habitat-Matterport 3D Dataset ({HM}3D): 1000 Large-scale 3D Environments for Embodied {AI}},
  author={Santhosh Kumar Ramakrishnan and Aaron Gokaslan and Erik Wijmans and Oleksandr Maksymets and Alexander Clegg and John M Turner and Eric Undersander and Wojciech Galuba and Andrew Westbury and Angel X Chang and Manolis Savva and Yili Zhao and Dhruv Batra},
  booktitle={Thirty-fifth Conference on Neural Information Processing Systems Datasets and Benchmarks Track},
  year={2021},
  url={https://arxiv.org/abs/2109.08238}
}

@inproceedings{huang2018deepmvs,
  title={Deepmvs: Learning multi-view stereopsis},
  author={Huang, Po-Han and Matzen, Kevin and Kopf, Johannes and Ahuja, Narendra and Huang, Jia-Bin},
  booktitle={Proceedings of the IEEE conference on computer vision and pattern recognition},
  pages={2821--2830},
  year={2018}
}

@inproceedings{dai2017scannet,
  title={Scannet: Richly-annotated 3d reconstructions of indoor scenes},
  author={Dai, Angela and Chang, Angel X and Savva, Manolis and Halber, Maciej and Funkhouser, Thomas and Nie{\ss}ner, Matthias},
  booktitle={Proceedings of the IEEE conference on computer vision and pattern recognition},
  pages={5828--5839},
  year={2017}
}

@inproceedings{yeshwanth2023scannet++,
  title={Scannet++: A high-fidelity dataset of 3d indoor scenes},
  author={Yeshwanth, Chandan and Liu, Yueh-Cheng and Nie{\ss}ner, Matthias and Dai, Angela},
  booktitle={Proceedings of the IEEE/CVF International Conference on Computer Vision},
  pages={12--22},
  year={2023}
}

@inproceedings{mehl2023spring,
  title={Spring: A high-resolution high-detail dataset and benchmark for scene flow, optical flow and stereo},
  author={Mehl, Lukas and Schmalfuss, Jenny and Jahedi, Azin and Nalivayko, Yaroslava and Bruhn, Andr{\'e}s},
  booktitle={Proceedings of the IEEE/CVF Conference on Computer Vision and Pattern Recognition},
  pages={4981--4991},
  year={2023}
}

@inproceedings{wang2020tartanair,
  title={Tartanair: A dataset to push the limits of visual slam},
  author={Wang, Wenshan and Zhu, Delong and Wang, Xiangwei and Hu, Yaoyu and Qiu, Yuheng and Wang, Chen and Hu, Yafei and Kapoor, Ashish and Scherer, Sebastian},
  booktitle={2020 IEEE/RSJ International Conference on Intelligent Robots and Systems (IROS)},
  pages={4909--4916},
  year={2020},
  organization={IEEE}
}

@article{bauer2019uasol,
  title={UASOL, a large-scale high-resolution outdoor stereo dataset},
  author={Bauer, Zuria and Gomez-Donoso, Francisco and Cruz, Edmanuel and Orts-Escolano, Sergio and Cazorla, Miguel},
  journal={Scientific data},
  volume={6},
  number={1},
  pages={162},
  year={2019},
  publisher={Nature Publishing Group UK London}
}

@inproceedings{tosi2021smd,
  title={Smd-nets: Stereo mixture density networks},
  author={Tosi, Fabio and Liao, Yiyi and Schmitt, Carolin and Geiger, Andreas},
  booktitle={Proceedings of the IEEE/CVF conference on computer vision and pattern recognition},
  pages={8942--8952},
  year={2021}
}

@article{cabon2020virtual,
  title={Virtual kitti 2},
  author={Cabon, Yohann and Murray, Naila and Humenberger, Martin},
  journal={arXiv preprint arXiv:2001.10773},
  year={2020}
}

@inproceedings{sun2020scalability,
  title={Scalability in perception for autonomous driving: Waymo open dataset},
  author={Sun, Pei and Kretzschmar, Henrik and Dotiwalla, Xerxes and Chouard, Aurelien and Patnaik, Vijaysai and Tsui, Paul and Guo, James and Zhou, Yin and Chai, Yuning and Caine, Benjamin and others},
  booktitle={Proceedings of the IEEE/CVF conference on computer vision and pattern recognition},
  pages={2446--2454},
  year={2020}
}

@inproceedings{xia2024rgbd,
  title={Rgbd objects in the wild: Scaling real-world 3d object learning from rgb-d videos},
  author={Xia, Hongchi and Fu, Yang and Liu, Sifei and Wang, Xiaolong},
  booktitle={Proceedings of the IEEE/CVF Conference on Computer Vision and Pattern Recognition},
  pages={22378--22389},
  year={2024}
}

@inproceedings{li2018megadepth,
  title={Megadepth: Learning single-view depth prediction from internet photos},
  author={Li, Zhengqi and Snavely, Noah},
  booktitle={Proceedings of the IEEE conference on computer vision and pattern recognition},
  pages={2041--2050},
  year={2018}
}

@inproceedings{wang2024dust3r,
  title={Dust3r: Geometric 3d vision made easy},
  author={Wang, Shuzhe and Leroy, Vincent and Cabon, Yohann and Chidlovskii, Boris and Revaud, Jerome},
  booktitle={Proceedings of the IEEE/CVF Conference on Computer Vision and Pattern Recognition},
  pages={20697--20709},
  year={2024}
}

@inproceedings{jang2025pow3r,
  title={Pow3r: Empowering unconstrained 3d reconstruction with camera and scene priors},
  author={Jang, Wonbong and Weinzaepfel, Philippe and Leroy, Vincent and Agapito, Lourdes and Revaud, Jerome},
  booktitle={Proceedings of the Computer Vision and Pattern Recognition Conference},
  pages={1071--1081},
  year={2025}
}

@inproceedings{wang2025cut3r,
  title={Continuous 3d perception model with persistent state},
  author={Wang, Qianqian and Zhang, Yifei and Holynski, Aleksander and Efros, Alexei A and Kanazawa, Angjoo},
  booktitle={Proceedings of the Computer Vision and Pattern Recognition Conference},
  pages={10510--10522},
  year={2025}
}

@inproceedings{leroy2024mast3r,
  title={Grounding image matching in 3d with mast3r},
  author={Leroy, Vincent and Cabon, Yohann and Revaud, J{\'e}r{\^o}me},
  booktitle={European Conference on Computer Vision},
  pages={71--91},
  year={2024},
  organization={Springer}
}

@inproceedings{hu2025depthcrafter,
  title={Depthcrafter: Generating consistent long depth sequences for open-world videos},
  author={Hu, Wenbo and Gao, Xiangjun and Li, Xiaoyu and Zhao, Sijie and Cun, Xiaodong and Zhang, Yong and Quan, Long and Shan, Ying},
  booktitle={Proceedings of the Computer Vision and Pattern Recognition Conference},
  pages={2005--2015},
  year={2025}
}

@inproceedings{butler2012naturalistic,
  title={A naturalistic open source movie for optical flow evaluation},
  author={Butler, Daniel J and Wulff, Jonas and Stanley, Garrett B and Black, Michael J},
  booktitle={European conference on computer vision},
  pages={611--625},
  year={2012},
  organization={Springer}
}

@misc{zhou2025omniworld,
      title={OmniWorld: A Multi-Domain and Multi-Modal Dataset for 4D World Modeling}, 
      author={Yang Zhou and Yifan Wang and Jianjun Zhou and Wenzheng Chang and Haoyu Guo and Zizun Li and Kaijing Ma and Xinyue Li and Yating Wang and Haoyi Zhu and Mingyu Liu and Dingning Liu and Jiange Yang and Zhoujie Fu and Junyi Chen and Chunhua Shen and Jiangmiao Pang and Kaipeng Zhang and Tong He},
      year={2025},
      eprint={2509.12201},
      archivePrefix={arXiv},
      primaryClass={cs.CV},
      url={https://arxiv.org/abs/2509.12201}, 
}

@inproceedings{palazzolo2019refusion,
  title={ReFusion: 3D reconstruction in dynamic environments for RGB-D cameras exploiting residuals},
  author={Palazzolo, Emanuele and Behley, Jens and Lottes, Philipp and Giguere, Philippe and Stachniss, Cyrill},
  booktitle={2019 IEEE/RSJ International Conference on Intelligent Robots and Systems (IROS)},
  pages={7855--7862},
  year={2019},
  organization={IEEE}
}

@inproceedings{silberman2012indoor,
  title={Indoor segmentation and support inference from rgbd images},
  author={Silberman, Nathan and Hoiem, Derek and Kohli, Pushmeet and Fergus, Rob},
  booktitle={European conference on computer vision},
  pages={746--760},
  year={2012},
  organization={Springer}
}

@article{zhang2024monst3r,
  title={Monst3r: A simple approach for estimating geometry in the presence of motion},
  author={Zhang, Junyi and Herrmann, Charles and Hur, Junhwa and Jampani, Varun and Darrell, Trevor and Cole, Forrester and Sun, Deqing and Yang, Ming-Hsuan},
  journal={arXiv preprint arXiv:2410.03825},
  year={2024}
}

@inproceedings{wang2025continuous,
  title={Continuous 3d perception model with persistent state},
  author={Wang, Qianqian and Zhang, Yifei and Holynski, Aleksander and Efros, Alexei A and Kanazawa, Angjoo},
  booktitle={Proceedings of the Computer Vision and Pattern Recognition Conference},
  pages={10510--10522},
  year={2025}
}

@inproceedings{schroppel2022benchmark,
  title={A benchmark and a baseline for robust multi-view depth estimation},
  author={Schr{\"o}ppel, Philipp and Bechtold, Jan and Amiranashvili, Artemij and Brox, Thomas},
  booktitle={2022 International Conference on 3D Vision (3DV)},
  pages={637--645},
  year={2022},
  organization={IEEE}
}

@inproceedings{schops2017multi,
  title={A multi-view stereo benchmark with high-resolution images and multi-camera videos},
  author={Schops, Thomas and Schonberger, Johannes L and Galliani, Silvano and Sattler, Torsten and Schindler, Konrad and Pollefeys, Marc and Geiger, Andreas},
  booktitle={Proceedings of the IEEE conference on computer vision and pattern recognition},
  pages={3260--3269},
  year={2017}
}

@article{aanaes2016large,
  title={Large-scale data for multiple-view stereopsis},
  author={Aan{\ae}s, Henrik and Jensen, Rasmus Ramsb{\o}l and Vogiatzis, George and Tola, Engin and Dahl, Anders Bjorholm},
  journal={International Journal of Computer Vision},
  volume={120},
  number={2},
  pages={153--168},
  year={2016},
  publisher={Springer}
}

@article{knapitsch2017tanks,
  title={Tanks and temples: Benchmarking large-scale scene reconstruction},
  author={Knapitsch, Arno and Park, Jaesik and Zhou, Qian-Yi and Koltun, Vladlen},
  journal={ACM Transactions on Graphics (ToG)},
  volume={36},
  number={4},
  pages={1--13},
  year={2017},
  publisher={ACM New York, NY, USA}
}

@inproceedings{schonberger2016structure,
  title={Structure-from-motion revisited},
  author={Schonberger, Johannes L and Frahm, Jan-Michael},
  booktitle={Proceedings of the IEEE conference on computer vision and pattern recognition},
  pages={4104--4113},
  year={2016}
}

@inproceedings{schonberger2016pixelwise,
  title={Pixelwise view selection for unstructured multi-view stereo},
  author={Sch{\"o}nberger, Johannes L and Zheng, Enliang and Frahm, Jan-Michael and Pollefeys, Marc},
  booktitle={European conference on computer vision},
  pages={501--518},
  year={2016},
  organization={Springer}
}

@inproceedings{yao2018mvsnet,
  title={Mvsnet: Depth inference for unstructured multi-view stereo},
  author={Yao, Yao and Luo, Zixin and Li, Shiwei and Fang, Tian and Quan, Long},
  booktitle={Proceedings of the European conference on computer vision (ECCV)},
  pages={767--783},
  year={2018}
}

@article{zhang2020visibility,
  title={Visibility-aware multi-view stereo network},
  author={Zhang, Jingyang and Yao, Yao and Li, Shiwei and Luo, Zixin and Fang, Tian},
  journal={arXiv preprint arXiv:2008.07928},
  year={2020}
}

@article{cao2024mvsformer++,
  title={Mvsformer++: Revealing the devil in transformer's details for multi-view stereo},
  author={Cao, Chenjie and Ren, Xinlin and Fu, Yanwei},
  journal={arXiv preprint arXiv:2401.11673},
  year={2024}
}

@inproceedings{reizenstein2021common,
  title={Common objects in 3d: Large-scale learning and evaluation of real-life 3d category reconstruction},
  author={Reizenstein, Jeremy and Shapovalov, Roman and Henzler, Philipp and Sbordone, Luca and Labatut, Patrick and Novotny, David},
  booktitle={Proceedings of the IEEE/CVF international conference on computer vision},
  pages={10901--10911},
  year={2021}
}

@article{zhou2018stereo,
  title={Stereo magnification: Learning view synthesis using multiplane images},
  author={Zhou, Tinghui and Tucker, Richard and Flynn, John and Fyffe, Graham and Snavely, Noah},
  journal={arXiv preprint arXiv:1805.09817},
  year={2018}
}

@inproceedings{sarlin2020superglue,
  title={Superglue: Learning feature matching with graph neural networks},
  author={Sarlin, Paul-Edouard and DeTone, Daniel and Malisiewicz, Tomasz and Rabinovich, Andrew},
  booktitle={Proceedings of the IEEE/CVF conference on computer vision and pattern recognition},
  pages={4938--4947},
  year={2020}
}

@inproceedings{lindenberger2021pixel,
  title={Pixel-perfect structure-from-motion with featuremetric refinement},
  author={Lindenberger, Philipp and Sarlin, Paul-Edouard and Larsson, Viktor and Pollefeys, Marc},
  booktitle={Proceedings of the IEEE/CVF international conference on computer vision},
  pages={5987--5997},
  year={2021}
}

@inproceedings{wang2023posediffusion,
  title={Posediffusion: Solving pose estimation via diffusion-aided bundle adjustment},
  author={Wang, Jianyuan and Rupprecht, Christian and Novotny, David},
  booktitle={Proceedings of the IEEE/CVF International Conference on Computer Vision},
  pages={9773--9783},
  year={2023}
}

@inproceedings{tang2025mv,
  title={Mv-dust3r+: Single-stage scene reconstruction from sparse views in 2 seconds},
  author={Tang, Zhenggang and Fan, Yuchen and Wang, Dilin and Xu, Hongyu and Ranjan, Rakesh and Schwing, Alexander and Yan, Zhicheng},
  booktitle={Proceedings of the Computer Vision and Pattern Recognition Conference},
  pages={5283--5293},
  year={2025}
}

@inproceedings{zhang2025flare,
  title={Flare: Feed-forward geometry, appearance and camera estimation from uncalibrated sparse views},
  author={Zhang, Shangzhan and Wang, Jianyuan and Xu, Yinghao and Xue, Nan and Rupprecht, Christian and Zhou, Xiaowei and Shen, Yujun and Wetzstein, Gordon},
  booktitle={Proceedings of the Computer Vision and Pattern Recognition Conference},
  pages={21936--21947},
  year={2025}
}

@inproceedings{yang2025fast3r,
  title={Fast3r: Towards 3d reconstruction of 1000+ images in one forward pass},
  author={Yang, Jianing and Sax, Alexander and Liang, Kevin J and Henaff, Mikael and Tang, Hao and Cao, Ang and Chai, Joyce and Meier, Franziska and Feiszli, Matt},
  booktitle={Proceedings of the Computer Vision and Pattern Recognition Conference},
  pages={21924--21935},
  year={2025}
}

@article{achiam2023gpt,
  title={Gpt-4 technical report},
  author={Achiam, Josh and Adler, Steven and Agarwal, Sandhini and Ahmad, Lama and Akkaya, Ilge and Aleman, Florencia Leoni and Almeida, Diogo and Altenschmidt, Janko and Altman, Sam and Anadkat, Shyamal and others},
  journal={arXiv preprint arXiv:2303.08774},
  year={2023}
}

@article{mildenhall2021nerf,
  title={Nerf: Representing scenes as neural radiance fields for view synthesis},
  author={Mildenhall, Ben and Srinivasan, Pratul P and Tancik, Matthew and Barron, Jonathan T and Ramamoorthi, Ravi and Ng, Ren},
  journal={Communications of the ACM},
  volume={65},
  number={1},
  pages={99--106},
  year={2021},
  publisher={ACM New York, NY, USA}
}

@article{kerbl20233dgs,
  title={3D Gaussian splatting for real-time radiance field rendering.},
  author={Kerbl, Bernhard and Kopanas, Georgios and Leimk{\"u}hler, Thomas and Drettakis, George},
  journal={ACM Trans. Graph.},
  volume={42},
  number={4},
  pages={139--1},
  year={2023}
}

@inproceedings{keetha2024splatamrb,
  title={Splatam: Splat track \& map 3d gaussians for dense rgb-d slam},
  author={Keetha, Nikhil and Karhade, Jay and Jatavallabhula, Krishna Murthy and Yang, Gengshan and Scherer, Sebastian and Ramanan, Deva and Luiten, Jonathon},
  booktitle={Proceedings of the IEEE/CVF Conference on Computer Vision and Pattern Recognition},
  pages={21357--21366},
  year={2024}
}

@article{hu2023towardrb,
  title={Toward general-purpose robots via foundation models: A survey and meta-analysis},
  author={Hu, Yafei and Xie, Quanting and Jain, Vidhi and Francis, Jonathan and Patrikar, Jay and Keetha, Nikhil and Kim, Seungchan and Xie, Yaqi and Zhang, Tianyi and Fang, Hao-Shu and others},
  journal={arXiv preprint arXiv:2312.08782},
  year={2023}
}

@inproceedings{Sun2020WaymoOpenDataset,
  author    = {Pei Sun and Henrik Kretzschmar and Xerxes Dotiwalla and Aurelien Chouard and Vijaysai Patnaik and Paul Tsui and James Guo and Yin Zhou and Yuning Chai and Benjamin Caine and Vijay Vasudevan and Wei Han and Jiquan Ngiam and Hang Zhao and Aleksei Timofeev and Scott Ettinger and Maxim Krivokon and Amy Gao and Aditya Joshi and Yu Zhang and Jonathon Shlens and Zhifeng Chen and Dragomir Anguelov},
  title     = {Scalability in Perception for Autonomous Driving: Waymo Open Dataset},
  booktitle = {Proceedings of the IEEE/CVF Conference on Computer Vision and Pattern Recognition (CVPR)},
  pages     = {2446--2454},
  month     = {June},
  year      = {2020}
}

@inproceedings{Caesar2020nuScenes,
  author    = {Holger Caesar and Varun Bankiti and Alex H. Lang and Sourabh Vora and Venice Erin Liong and Qiang Xu and Anush Krishnan and Yu Pan and Giancarlo Baldan and Oscar Beijbom},
  title     = {nuScenes: A Multimodal Dataset for Autonomous Driving},
  booktitle = {Proceedings of the IEEE/CVF Conference on Computer Vision and Pattern Recognition (CVPR)},
  pages     = {11621--11631},
  month     = {June},
  year      = {2020}
}

@inproceedings{Lang2019PointPillars,
  author    = {Alex H. Lang and Sourabh Vora and Holger Caesar and Lubing Zhou and Jiong Yang and Oscar Beijbom},
  title     = {PointPillars: Fast Encoders for Object Detection From Point Clouds},
  booktitle = {Proceedings of the IEEE/CVF Conference on Computer Vision and Pattern Recognition (CVPR)},
  pages     = {12697--12705},
  month     = {June},
  year      = {2019}
}

@article{Dai2017BundleFusion,
  author    = {Angela Dai and Matthias Nie{\ss}ner and Michael Zollh{\"o}fer and Shahram Izadi and Christian Theobalt},
  title     = {BundleFusion: Real-Time Globally Consistent 3D Reconstruction Using On-the-Fly Surface Reintegration},
  journal   = {ACM Transactions on Graphics (Proc. SIGGRAPH)},
  volume    = {36},
  number    = {4},
  year      = {2017},
  doi       = {10.1145/3072959.3054739}
}

@inproceedings{Peng2024RTGSLAM,
  author    = {Zhexi Peng and Tianjia Shao and Yong Liu and Jingke Zhou and Yin Yang and Jingdong Wang and Kun Zhou},
  title     = {RTG-SLAM: Real-time 3D Reconstruction at Scale using Gaussian Splatting},
  booktitle = {ACM SIGGRAPH 2024 Conference Papers},
  year      = {2024},
  note      = {Also available as arXiv:2404.19706}
}

@inproceedings{Hu2024CGSLAM,
  author    = {Jiarui Hu and Xianhao Chen and Boyin Feng and Guanglin Li and Liangjing Yang and Hujun Bao and Guofeng Zhang and Zhaopeng Cui},
  title     = {CG-SLAM: Efficient Dense RGB-D SLAM in a Consistent Uncertainty-Aware 3D Gaussian Field},
  booktitle = {Computer Vision -- ECCV 2024, Lecture Notes in Computer Science},
  volume    = {15083},
  pages     = {93--112},
  year      = {2024},
  publisher = {Springer}
}

@book{hartley2003multiple,
  title={Multiple view geometry in computer vision},
  author={Hartley, Richard and Zisserman, Andrew},
  year={2003},
  publisher={Cambridge university press}
}

@article{wang20243d,
  title={3d reconstruction with spatial memory},
  author={Wang, Hengyi and Agapito, Lourdes},
  journal={arXiv preprint arXiv:2408.16061},
  year={2024}
}

@inproceedings{shotton2013scene,
  title={Scene coordinate regression forests for camera relocalization in RGB-D images},
  author={Shotton, Jamie and Glocker, Ben and Zach, Christopher and Izadi, Shahram and Criminisi, Antonio and Fitzgibbon, Andrew},
  booktitle={Proceedings of the IEEE conference on computer vision and pattern recognition},
  pages={2930--2937},
  year={2013}
}

@inproceedings{liu2025slam3r,
  title={Slam3r: Real-time dense scene reconstruction from monocular rgb videos},
  author={Liu, Yuzheng and Dong, Siyan and Wang, Shuzhe and Yin, Yingda and Yang, Yanchao and Fan, Qingnan and Chen, Baoquan},
  booktitle={Proceedings of the Computer Vision and Pattern Recognition Conference},
  pages={16651--16662},
  year={2025}
}

@inproceedings{azinovic2022neural,
  title={Neural rgb-d surface reconstruction},
  author={Azinovi{\'c}, Dejan and Martin-Brualla, Ricardo and Goldman, Dan B and Nie{\ss}ner, Matthias and Thies, Justus},
  booktitle={Proceedings of the IEEE/CVF Conference on Computer Vision and Pattern Recognition},
  pages={6290--6301},
  year={2022}
}

@ARTICLE{3dgssurvey2025,
  author={Bao, Yanqi and Ding, Tianyu and Huo, Jing and Liu, Yaoli and Li, Yuxin and Li, Wenbin and Gao, Yang and Luo, Jiebo},
  journal={IEEE Transactions on Circuits and Systems for Video Technology}, 
  title={3D Gaussian Splatting: Survey, Technologies, Challenges, and Opportunities}, 
  year={2025},
  volume={35},
  number={7},
  pages={6832-6852},
  doi={10.1109/TCSVT.2025.3538684}}

@ARTICLE{3dgssurvey2024,
  author={Fei, Ben and Xu, Jingyi and Zhang, Rui and Zhou, Qingyuan and Yang, Weidong and He, Ying},
  journal={IEEE Transactions on Visualization and Computer Graphics}, 
  title={3D Gaussian Splatting as a New Era: A Survey}, 
  year={2025},
  volume={31},
  number={8},
  pages={4429-4449},
  doi={10.1109/TVCG.2024.3397828}}

@article{gao2022nerf,
  title={Nerf: Neural radiance field in 3d vision, a comprehensive review},
  author={Gao, Kyle and Gao, Yina and He, Hongjie and Lu, Dening and Xu, Linlin and Li, Jonathan},
  journal={arXiv preprint arXiv:2210.00379},
  year={2022}
}

@article{wang2024nerfs,
  title={NeRFs in robotics: A survey},
  author={Wang, Guangming and Pan, Lei and Peng, Songyou and Liu, Shaohui and Xu, Chenfeng and Miao, Yanzi and Zhan, Wei and Tomizuka, Masayoshi and Pollefeys, Marc and Wang, Hesheng},
  journal={The International Journal of Robotics Research},
  pages={02783649251374246},
  year={2024},
  publisher={SAGE Publications Sage UK: London, England}
}

@article{lan2025stream3r,
  title={STream3R: Scalable Sequential 3D Reconstruction with Causal Transformer},
  author={Lan, Yushi and Luo, Yihang and Hong, Fangzhou and Zhou, Shangchen and Chen, Honghua and Lyu, Zhaoyang and Yang, Shuai and Dai, Bo and Loy, Chen Change and Pan, Xingang},
  journal={arXiv preprint arXiv:2508.10893},
  year={2025}
}

@article{zhuo2025streaming,
  title={Streaming 4d visual geometry transformer},
  author={Zhuo, Dong and Zheng, Wenzhao and Guo, Jiahe and Wu, Yuqi and Zhou, Jie and Lu, Jiwen},
  journal={arXiv preprint arXiv:2507.11539},
  year={2025}
}

@article{maggio2025vggt,
  title={Vggt-slam: Dense rgb slam optimized on the sl (4) manifold},
  author={Maggio, Dominic and Lim, Hyungtae and Carlone, Luca},
  journal={arXiv preprint arXiv:2505.12549},
  year={2025}
}

@article{liu2025vggt,
  title={VGGT-X: When VGGT Meets Dense Novel View Synthesis},
  author={Liu, Yang and Luo, Chuanchen and Tang, Zimo and Peng, Junran and Zhang, Zhaoxiang},
  journal={arXiv preprint arXiv:2509.25191},
  year={2025}
}

@article{peng2023kosmos,
  title={Kosmos-2: Grounding Multimodal Large Language Models to the World},
  author={Peng, Zhiliang and Wang, Wenhui and Dong, Li and Hao, Yaru and Huang, Shaohan and Ma, Shuming and Wei, Furu},
  journal={arXiv preprint arXiv:2306.14824},
  year={2023}
}

@article{li2023generalist,
    title={Towards Generalist Robot Policies: What Matters in Building Vision-Language-Action Models},
    author={Li, Xinghang and Li, Peiyan and Liu, Minghuan and Wang, Dong and Liu, Jirong and Kang, Bingyi and Ma, Xiao and Kong, Tao and Zhang, Hanbo and Liu, Huaping},
    journal={arXiv preprint arXiv:2412.14058},
    year={2024}
}

@misc{ze2025manipulation3d,
      title={Generalizable Humanoid Manipulation with 3D Diffusion Policies}, 
      author={Yanjie Ze and Zixuan Chen and Wenhao Wang and Tianyi Chen and Xialin He and Ying Yuan and Xue Bin Peng and Jiajun Wu},
      year={2025},
      eprint={2410.10803},
      archivePrefix={arXiv},
      primaryClass={cs.RO},
      url={https://arxiv.org/abs/2410.10803}, 
}

@article{mees2022calvin,
author = {Oier Mees and Lukas Hermann and Erick Rosete-Beas and Wolfram Burgard},
title = {CALVIN: A Benchmark for Language-Conditioned Policy Learning for Long-Horizon Robot Manipulation Tasks},
journal={IEEE Robotics and Automation Letters (RA-L)},
volume={7},
number={3},
pages={7327-7334},
year={2022}
}
}
\clearpage
\setcounter{page}{1}
\maketitlesupplementary
\appendix
\section{Appendix}
\subsection{Dataset Details}
We train our model on 19 datasets that contain a diverse range of scene types, including: 
ARKitScenes~\cite{baruch2021arkitscenes}, BlendedMVS~\cite{yao2020blendedmvs}, DL3DV~\cite{ling2024dl3dv}, Dynamic Replica~\cite{karaev2023dynamicstereo}, HyperSim~\cite{roberts2021hypersim}, Kubric~\cite{greff2022kubric}, MapFree~\cite{arnold2022map}, MegaDepth~\cite{li2018megadepth}, Matterport 3D~\cite{ramakrishnan2021hm3d}, MVS-Synth~\cite{huang2018deepmvs}, ScanNet~\cite{dai2017scannet}, ScanNet++~\cite{yeshwanth2023scannet++}, Spring~\cite{mehl2023spring}, TartanAir~\cite{wang2020tartanair}, UASOL~\cite{bauer2019uasol}, Unreal 4K~\cite{tosi2021smd}, Virtual KITTI~\cite{cabon2020virtual}, Waymo~\cite{sun2020scalability}, WildRGBD~\cite{xia2024rgbd}. 
We modified the official DUSt3R~\cite{wang2024dust3r} dataloader script to adapt it for the VGGT~\cite{wang2025vggt} training process. 
Table~\ref{dataset-table} summarizes the statistics of the datasets we used. 
During training, in each epoch, we sample a fixed total number of samples from the 
training datasets, with their proportions indicated by the ``ratio'' column in the table.
Note that the number of images may differ from the full official release. 

\begin{table*}[htbp]

\caption{\textbf{Training datasets statistic.} Each training epoch is composed of 20 datasets, and their relative quantities are reported as a ratio.}
\label{dataset-table}
\renewcommand{\arraystretch}{1.1}
\centering
\resizebox{0.8\textwidth}{!}{%
\begin{tabular}{@{}c|c|c|c|c|c|c@{}}
\toprule
\textbf{Index} & \textbf{Dataset} & \textbf{Scene Type} & \textbf{Real/Synthetic} & \textbf{Dynamic} & \textbf{\# of Frames} & \textbf{Training Prob (\%)} \\ \midrule
1 & \textbf{ARKitScene}~\cite{baruch2021arkitscenes} & Indoor & Real & Static & 1.2M & 2.07 \\
2 & \textbf{BlendedMVS}~\cite{yao2020blendedmvs} & Mixed & Real & Static & 1.1M & 2.07 \\
3 & \textbf{DL3DV}~\cite{ling2024dl3dv} & Mixed & Real & Static & 21M & 17.81 \\
4 & \textbf{Dynamic Replica}~\cite{karaev2023dynamicstereo} & Indoor & Synthetic & Dynamic & 2.8M & 4.68 \\
5 & \textbf{HyperSim}~\cite{roberts2021hypersim} & Indoor & Synthetic & Static & 70K & 3.55 \\
6 & \textbf{Kubric}~\cite{greff2022kubric} & Object & Synthetic & Dynamic & 1.3M & 1.67 \\
7 & \textbf{MapFree}~\cite{arnold2022map} & Outdoor & Real & Static & 2.6M & 9.03 \\
8 & \textbf{MegaDepth}~\cite{li2018megadepth} & Outdoor & Real & Static & 1.2M & 2.07 \\
9 & \textbf{Matterport 3D}~\cite{ramakrishnan2021hm3d} & Indoor & Real & Static & 1.9M & 3.48 \\
10 & \textbf{MVS-Synth}~\cite{huang2018deepmvs} & Outdoor & Synthetic & Static & 12K & 1.34 \\
11 & \textbf{ScanNet}~\cite{dai2017scannet} & Indoor & Real & Static & 23M & 4.75 \\
12 & \textbf{ScanNet++}~\cite{yeshwanth2023scannet++} & Indoor & Real & Static & 7.8M & 13.88 \\
13 & \textbf{Spring}~\cite{mehl2023spring} & Outdoor & Synthetic & Dynamic & 4.9K & 0.5 \\
14 & \textbf{Tartanair}~\cite{wang2020tartanair} & Mixed & Synthetic & Static & 3M & 9.36 \\
15 & \textbf{Uasol}~\cite{bauer2019uasol} & Outdoor & Real & Static & 1.3M & 2.41 \\
16 & \textbf{Unreal 4K}~\cite{tosi2021smd} & Outdoor & Synthetic & Static & 16K & 1.81 \\
17 & \textbf{Vkitti}~\cite{cabon2020virtual} & Outdoor & Synthetic & Dynamic & 42K & 3.44 \\
18 & \textbf{Waymo}~\cite{sun2020scalability} & Outdoor & Real & Dynamic & 7.9M & 8.36 \\
19 & \textbf{WildRGBD}~\cite{xia2024rgbd} & Object & Real & Static & 1.9M & 4.35 \\
\bottomrule
\end{tabular}}
\end{table*}

\subsection{More Implementation details}
\textbf{Training Objective.}
In OmniVGGT, all input images, together with the available camera parameters and depth maps (if provided), are fed into the network $\mathcal{G}$, which predicts in an end-to-end manner the 3D point maps, complete camera poses, intrinsics, depth maps, and their corresponding confidence maps:
\begin{equation}
\mathcal{G}\left(\mathbf{I},\mathbf{C},\mathbf{D}\right)=(\hat{C_i},\hat{P_i},\hat{D_i}, \hat{Y_i})_{i=1}^N
\end{equation}
The training objectives of \ours consist of three components: camera, depth, and point map.
\begin{equation}
\mathcal{L}=\mathcal{L}_{\text {camera }}+\mathcal{L}_{\text {depth }}+\mathcal{L}_{\text {pmap }},
\end{equation}
The camera loss $\mathcal{L}$ camera supervises the cameras $\hat{\mathbf{g}}_i$ with ground truth $\mathbf{g}_i$ using L1 loss:
\begin{equation}
\mathcal{L}_{\text {camera }}=\sum_{i=1}^N\left\|\hat{\mathbf{g}}_i-\mathbf{g}_i\right\|_1
\end{equation}
Following VGGT, we apply a confidence-aware regression loss to the depth and point map, both along with a gradient-based term:
\begin{equation}
\begin{split}
\mathcal{L}_{\mathrm{depth}}
  &= \sum_{i=1}^N \left\|
     \Sigma_i^D \odot \bigl(\hat{D}_i - D_i\bigr) \right\|
     + \left\| \Sigma_i^D \odot
     \bigl(\nabla \hat{D}_i - \nabla D_i\bigr) \right\| \\
  &\quad - \alpha \log \Sigma_i^D
\end{split}
\end{equation}
\begin{equation}
\begin{split}
\mathcal{L}_{\mathrm{pmap}}
  &= \sum_{i=1}^N \left\|
     \Sigma_i^P \odot \bigl(\hat{P}_i - P_i\bigr) \right\|
     + \left\| \Sigma_i^P \odot
     \bigl(\nabla \hat{P}_i - \nabla P_i\bigr) \right\| \\
  &\quad - \alpha \log \Sigma_i^P
\end{split}
\end{equation}
where $\odot$ is the channel-broadcast element-wise product and $\alpha$ is a hyper-parameter.
In addition, our depth $D$, point map $P$, and camera translations $t$ in ground truths are all normalized by dividing the average Euclidean distance of all 3D points in the point map $P$ to the origin.
It should be noted that this normalization process is different from the normalization used in the information injection stage of \ours.

\noindent \textbf{Frame Sampling Strategy.} For every batch, we select between 2 and 24 frames from multiple random training scenes while maintaining a constant total of 24 frames within each batch.
We sample each batch of images based on camera pose similarity. 
For each frame, all other frames are ranked according to their pose similarity, and the top N most similar frames are selected as its valid range. 
Then, for each sequence, we randomly choose one frame as the anchor frame and sample the remaining frames from its valid range.
In addition, our depth $D$, point map $P$, and camera translations $t$ in ground truths are all normalized by dividing the average Euclidean distance of all 3D points in the point map $P$ to the origin.
It should be noted that this normalization process is different from the normalization used in the information injection stage.

\noindent \textbf{Training Details.}
We initialize \ours by using pre-trained weights from VGGT and fine-tune for 10 epochs of 12M iterations each.
We train the model using the AdamW optimizer with a learning rate of $2\times10^{-5}$ for prediction heads and $1\times10^{-5}$ for backbone, which incorporates a 5K step linear warmup and cosine weight decay schedule.
The $p$ in our training objective is set to $10\%$.
The spatial resolutions of the input images, depth maps, and point maps 
range from $(518\times168)$ to $(518\times518)$.
We also use ColorJitter as the data augmentation to enhance the model’s robustness in varying lighting conditions.

\subsection{Baselines.}
We mainly compare our approach with VGGT, Pow3R~\cite{jang2025pow3r}, and DUSt3R~\cite{wang2024dust3r}. 
Both Pow3R and DUSt3R are feed-forward models that take a pair of views as input.
The distinctive feature of Pow3R is that it can additionally incorporate camera intrinsics, poses, and depth maps as auxiliary inputs, but only in a pairwise input.

\subsection{More Results of Auxiliary Information Guidance}\label{more_main}

In this subsection, we present the performance of auxiliary information guidance on additional datasets. 
Tables~\ref{tab:maintable_arkitscene} and \ref{tab:maintable_omniworld} report the results on ARKitScene~\cite{baruch2021arkitscenes} and OmniWorld (Game)~\cite{zhou2025omniworld}, respectively.
The methods have not been trained on these datasets.

\begin{table*}[t]
\caption{\textbf{Impact of Auxiliary Information Injection on ARKitScenes (unseen)~\cite{baruch2021arkitscenes}.} We also show in green the absolute improvement w.r.t. the results without auxiliary information. The best results in each category are \textbf{bold}.}
\label{tab:maintable_arkitscene}
\centering
\resizebox{0.8\linewidth}{!}{
\begin{tabular}{@{}c|cc|cc|ccc@{}}
\toprule
\textbf{Method} & \multicolumn{2}{c|}{\textbf{Aux. information (\%)}} & \multicolumn{2}{c|}{\textbf{Depth}} & \multicolumn{3}{c}{\textbf{Camera}} \\ \cmidrule(l){2-8} 
\textbf{} & \multicolumn{1}{c|}{\textbf{Depth}} & \textbf{Camera} & \multicolumn{1}{c|}{\textbf{Abs Rel$\downarrow$}} & \textbf{$\delta<1.25\uparrow$} & \multicolumn{1}{c|}{\textbf{RRA@$5^{\circ}$}$\uparrow$} & \multicolumn{1}{c|}{\textbf{RTA@ $5^{\circ}$}$\uparrow$} & \textbf{AUC@$30^{\circ}\uparrow$} \\ \midrule
VGGT & \multicolumn{1}{c|}{\XSolidBrush} & \XSolidBrush & \multicolumn{1}{c|}{0.048} & 98.90 & \multicolumn{1}{c|}{72.17} & \multicolumn{1}{c|}{35.94} & 60.52 \\ 
\textbf{\ours} & \multicolumn{1}{c|}{\XSolidBrush} & \XSolidBrush & \multicolumn{1}{c|}{\textbf{0.035}} & \textbf{99.02} & \multicolumn{1}{c|}{\textbf{73.66}} & \multicolumn{1}{c|}{\textbf{50.03}} & \textbf{65.82} \\ \midrule
\multirow{10}{*}{\thead{\textbf{\ours +} \\ \textbf{aux. information}}} & \multicolumn{1}{c|}{30} & \XSolidBrush & \multicolumn{1}{c|}{0.035 \textcolor{teal}{(+0.000)}} & 99.57 \textcolor{teal}{(+0.55)} & \multicolumn{1}{c|}{77.53 \textcolor{teal}{(+3.87)}} & \multicolumn{1}{c|}{50.59 \textcolor{teal}{(+0.56)}} & 67.92 \textcolor{teal}{(+2.10)} \\
 & \multicolumn{1}{c|}{50} & \XSolidBrush & \multicolumn{1}{c|}{0.029 \textcolor{teal}{(+0.006)}} & 99.52 \textcolor{teal}{(+0.50)} & \multicolumn{1}{c|}{77.62 \textcolor{teal}{(+3.96)}} & \multicolumn{1}{c|}{51.97 \textcolor{teal}{(+1.94)}} & \textbf{68.50 \textcolor{teal}{(+2.68)}} \\
 & \multicolumn{1}{c|}{70} & \XSolidBrush & \multicolumn{1}{c|}{0.021 \textcolor{teal}{(+0.014)}} & 99.57 \textcolor{teal}{(+0.55)} & \multicolumn{1}{c|}{\textbf{76.34 \textcolor{teal}{(+2.68)}}} & \multicolumn{1}{c|}{53.15 \textcolor{teal}{(+3.12)}} &68.25 \textcolor{teal}{(+2.43)} \\
 & \multicolumn{1}{c|}{100} & \XSolidBrush & \multicolumn{1}{c|}{\textbf{0.006 \textcolor{teal}{(+0.029)}}} & \textbf{99.97 \textcolor{teal}{(+0.95)}} & \multicolumn{1}{c|}{74.59 \textcolor{teal}{(+0.93)}} & \multicolumn{1}{c|}{\textbf{53.69 \textcolor{teal}{(+3.66)}}} & 67.99 \textcolor{teal}{(+2.17)} \\ \cmidrule(l){2-8}
 & \multicolumn{1}{c|}{\XSolidBrush} & 30 & \multicolumn{1}{c|}{0.035 \textcolor{teal}{(+0.000)}} & 99.31 \textcolor{teal}{(+0.29)} & \multicolumn{1}{c|}{77.60 \textcolor{teal}{(+3.94)}} & \multicolumn{1}{c|}{51.40 \textcolor{teal}{(+1.37)}} & 68.46 \textcolor{teal}{(+2.64)} \\
 & \multicolumn{1}{c|}{\XSolidBrush} & 50 & \multicolumn{1}{c|}{0.035 \textcolor{teal}{(+0.000)}} & 99.30 \textcolor{teal}{(+0.28)} & \multicolumn{1}{c|}{79.69 \textcolor{teal}{(+6.03)}} & \multicolumn{1}{c|}{53.92 \textcolor{teal}{(+3.89)}} & 70.69 \textcolor{teal}{(+4.87)} \\
 & \multicolumn{1}{c|}{\XSolidBrush} & 70 & \multicolumn{1}{c|}{0.034 \textcolor{teal}{(+0.001)}} & 99.28 \textcolor{teal}{(+0.26)} & \multicolumn{1}{c|}{84.69 \textcolor{teal}{(+11.03)}} & \multicolumn{1}{c|}{56.59 \textcolor{teal}{(+6.56)}} & 74.21 \textcolor{teal}{(+8.39)} \\
 & \multicolumn{1}{c|}{\XSolidBrush} & 100 & \multicolumn{1}{c|}{\textbf{0.034 \textcolor{teal}{(+0.001)}}} & \textbf{99.89 \textcolor{teal}{(+0.87)}} & \multicolumn{1}{c|}{\textbf{91.20 \textcolor{teal}{(+17.54)}}} & \multicolumn{1}{c|}{\textbf{64.98 \textcolor{teal}{(+14.95)}}} & \textbf{81.64 \textcolor{teal}{(+15.82)}} \\ \cmidrule(l){2-8}
 & \multicolumn{1}{c|}{100} & 100 & \multicolumn{1}{c|}{\textbf{0.006 \textcolor{teal}{(+0.029)}}} & \textbf{99.97 \textcolor{teal}{(+0.95)}} & \multicolumn{1}{c|}{\textbf{90.19 \textcolor{teal}{(+16.53)}}} & \multicolumn{1}{c|}{\textbf{67.00 \textcolor{teal}{(+16.97)}}} & \textbf{81.91 \textcolor{teal}{(+16.09)}} \\
\bottomrule
\end{tabular}}
\end{table*}

\begin{table*}[t]
\caption{\textbf{Impact of Auxiliary Information Injection on OmniWorld-Game (unseen)~\cite{zhou2025omniworld}.}  We also show in green the absolute improvement w.r.t. the results without auxiliary information. The best results in each category are \textbf{bold}.}
\label{tab:maintable_omniworld}
\centering
\resizebox{0.8\linewidth}{!}{
\begin{tabular}{@{}c|cc|cc|ccc@{}}
\toprule
\textbf{Method} & \multicolumn{2}{c|}{\textbf{Aux. information (\%)}} & \multicolumn{2}{c|}{\textbf{Depth}} & \multicolumn{3}{c}{\textbf{Camera}} \\ \cmidrule(l){2-8} 
\textbf{} & \multicolumn{1}{c|}{\textbf{Depth}} & \textbf{Camera} & \multicolumn{1}{c|}{\textbf{Abs Rel$\downarrow$}} & \textbf{$\delta<1.25\uparrow$} & \multicolumn{1}{c|}{\textbf{RRA@$5^{\circ}$}$\uparrow$} & \multicolumn{1}{c|}{\textbf{RTA@ $5^{\circ}$}$\uparrow$} & \textbf{AUC@$30^{\circ}\uparrow$} \\ \midrule
VGGT~\cite{wang2025vggt} & \multicolumn{1}{c|}{\XSolidBrush} & \XSolidBrush & \multicolumn{1}{c|}{0.260} & 79.10 & \multicolumn{1}{c|}{73.97} & \multicolumn{1}{c|}{58.57} & 63.75 \\
\textbf{\ours} & \multicolumn{1}{c|}{\XSolidBrush} & \XSolidBrush & \multicolumn{1}{c|}{\textbf{0.240}} & \textbf{80.50} & \multicolumn{1}{c|}{\textbf{74.20}} & \multicolumn{1}{c|}{\textbf{59.39}} & \textbf{63.86} \\ \midrule
\multirow{10}{*}{\thead{\textbf{\ours +} \\ \textbf{aux. information}}} & \multicolumn{1}{c|}{30} & \XSolidBrush & \multicolumn{1}{c|}{0.208 \textcolor{teal}{(+0.032)}} & 87.28 \textcolor{teal}{(+6.78)} & \multicolumn{1}{c|}{74.42 \textcolor{teal}{(+0.22)}} & \multicolumn{1}{c|}{59.57 \textcolor{teal}{(+0.18)}} & 63.91 \textcolor{teal}{(+0.05)} \\
 & \multicolumn{1}{c|}{50} & \XSolidBrush & \multicolumn{1}{c|}{0.178 \textcolor{teal}{(+0.062)}} & 90.68 \textcolor{teal}{(+10.18)} & \multicolumn{1}{c|}{74.38 \textcolor{teal}{(+0.18)}} & \multicolumn{1}{c|}{59.48 \textcolor{teal}{(+0.09)}} & 63.98 \textcolor{teal}{(+0.12)} \\
 & \multicolumn{1}{c|}{70} & \XSolidBrush & \multicolumn{1}{c|}{0.151 \textcolor{teal}{(+0.089)}} & 92.23 \textcolor{teal}{(+11.73)} & \multicolumn{1}{c|}{\textbf{74.63 \textcolor{teal}{(+0.43)}}} & \multicolumn{1}{c|}{59.82 \textcolor{teal}{(+0.43)}} & 64.20 \textcolor{teal}{(+0.34)} \\
 & \multicolumn{1}{c|}{100} & \XSolidBrush & \multicolumn{1}{c|}{\textbf{0.095 \textcolor{teal}{(+0.145)}}} & \textbf{94.54 \textcolor{teal}{(+14.04)}} & \multicolumn{1}{c|}{74.31 \textcolor{teal}{(+0.11)}} & \multicolumn{1}{c|}{\textbf{60.54 \textcolor{teal}{(+1.15)}}} & \textbf{64.87 \textcolor{teal}{(+1.01)}} \\ \cmidrule(l){2-8}
 & \multicolumn{1}{c|}{\XSolidBrush} & 30 & \multicolumn{1}{c|}{0.239 \textcolor{teal}{(+0.001)}} & 80.50 \textcolor{teal}{(+0.00)} & \multicolumn{1}{c|}{74.29 \textcolor{teal}{(+0.09)}} & \multicolumn{1}{c|}{60.70 \textcolor{teal}{(+1.31)}} & 64.75 \textcolor{teal}{(+0.89)} \\
 & \multicolumn{1}{c|}{\XSolidBrush} & 50 & \multicolumn{1}{c|}{0.238 \textcolor{teal}{(+0.002)}} & 80.51 \textcolor{teal}{(+0.01)} & \multicolumn{1}{c|}{74.37 \textcolor{teal}{(+0.17)}} & \multicolumn{1}{c|}{61.93 \textcolor{teal}{(+2.54)}} & 65.48 \textcolor{teal}{(+1.62)} \\
 & \multicolumn{1}{c|}{\XSolidBrush} & 70 & \multicolumn{1}{c|}{0.237 \textcolor{teal}{(+0.003)}} & 80.51 \textcolor{teal}{(+0.01)} & \multicolumn{1}{c|}{74.42 \textcolor{teal}{(+0.22)}} & \multicolumn{1}{c|}{64.15 \textcolor{teal}{(+4.76)}} & 68.06 \textcolor{teal}{(+4.20)} \\
 & \multicolumn{1}{c|}{\XSolidBrush} & 100 & \multicolumn{1}{c|}{\textbf{0.237 \textcolor{teal}{(+0.003)}}} & \textbf{80.52 \textcolor{teal}{(+0.02)}} & \multicolumn{1}{c|}{\textbf{77.91 \textcolor{teal}{(+3.71)}}} & \multicolumn{1}{c|}{\textbf{65.46 \textcolor{teal}{(+6.07)}}} & \textbf{71.91 \textcolor{teal}{(+8.05)}} \\ \cmidrule(l){2-8}
 & \multicolumn{1}{c|}{100} & 100 & \multicolumn{1}{c|}{\textbf{0.094 \textcolor{teal}{(+0.146)}}} & \textbf{94.57 \textcolor{teal}{(+14.07)}} & \multicolumn{1}{c|}{\textbf{79.43 \textcolor{teal}{(+5.23)}}} & \multicolumn{1}{c|}{\textbf{68.89 \textcolor{teal}{(+9.50)}}} & \textbf{74.35 \textcolor{teal}{(+10.49)}} \\
\bottomrule
\end{tabular}}
\end{table*}

\subsection{Full Results of Multi-View Depth Estimation}
In this subsection, we present the complete results of the multi-view depth evaluation in Section~\ref{sec:depth}. 
For \ours, all images are resized to a fixed width of 518 pixels, and the aspect ratio is adjusted to the closest aspect ratio used during training according to the original images.
The reported results are averaged over all samples.
As shown in Table~\ref{tab:fullmultiviewDepth}, we further analyze the impact of injecting different percentages of depth information.
We observe that both VGGT and \ours show relatively poor Rel performance on the ScanNet dataset, mainly because the ground-truth depth in ScanNet is noisy (e.g., walls and floors are not smooth).
Although the predicted depth maps are visually reasonable, the quantitative metrics appear degraded.

\begin{table*}[t]
\caption{\textbf{Multi-view Depth Evaluation.}  (Parentheses) denote training on data from the same domain. K”, “RT”, and “D” denote intrinsic, relative pose, and depth information, respectively. 
The best and second best results are \textbf{bold} and \underline{underlined} respectively.}
\label{tab:fullmultiviewDepth}
\centering
\footnotesize
\resizebox{0.9\linewidth}{!}{
\begin{tabular}{@{}l|cc|cccccccccc@{}}
\toprule
\multicolumn{1}{c|}{\multirow{2}{*}{\textbf{Method}}} & \multicolumn{1}{c|}{\textbf{GT}} & \multicolumn{1}{c|}{\textbf{Extra}} & \multicolumn{2}{c}{\textbf{ScanNet~\cite{dai2017scannet}}} & \multicolumn{2}{c}{\textbf{ETH3D~\cite{schops2017multi}}} & \multicolumn{2}{c}{\textbf{DTU~\cite{aanaes2016large}}} & \multicolumn{2}{c}{\textbf{T\&T~\cite{knapitsch2017tanks}}} & \multicolumn{2}{|c}{\textbf{Average}} \\ \cmidrule(l){4-13} 
 & \multicolumn{1}{c|}{\textbf{range}} & \multicolumn{1}{c|}{\textbf{info}} & \textbf{rel$\downarrow$} & \textbf{$\tau\uparrow$} & \textbf{rel$\downarrow$} & \textbf{$\tau\uparrow$} & \textbf{rel$\downarrow$} & \textbf{$\tau\uparrow$} & \textbf{rel$\downarrow$} & \multicolumn{1}{c|}{\textbf{$\tau\uparrow$}} & \textbf{rel$\downarrow$} & \textbf{$\tau\uparrow$} \\ \midrule
COLMAP~\cite{schonberger2016pixelwise, schonberger2016structure} (K+RT) & \multicolumn{1}{c|}{$\times$} & $\times$ & 14.6 & 34.2 & 16.4 & 55.1 & 0.7 & 96.5 & 2.7 & \multicolumn{1}{c|}{95.0} & 8.6 & 70.2 \\
COLMAP Dense~\cite{schonberger2016pixelwise, schonberger2016structure} K+RT & \multicolumn{1}{c|}{$\times$} & $\times$ & 38.0 & 22.5 & 89.8 & 23.2 & 20.8 & 69.3 & 25.7 & \multicolumn{1}{c|}{76.4} & 43.6 & 47.9 \\
MVSNet~\cite{yao2018mvsnet} (K+RT) & \multicolumn{1}{c|}{$\checkmark$} & $\times$ & 22.7 & 20.9 & 21.6 & 35.6 & (1.8) & (86.7) & 6.5 & \multicolumn{1}{c|}{74.6} & 12.3 & 11.1 \\
Vis-MVSNet~\cite{zhang2020visibility} (K+RT) & \multicolumn{1}{c|}{$\checkmark$} & $\times$ & 8.9 & 33.5 & 10.8 & 43.3 & 1.8 & 87.4 & 4.1 & \multicolumn{1}{c|}{7.2} & 6.4 & 42.9 \\
MVS-Former++ (K+RT) & \multicolumn{1}{c|}{$\checkmark$} & $\times$ & 15.2 & 21.9 & 21.4 & 32.5 & (1.2) & (91.9) & 7.6 & \multicolumn{1}{c|}{71.5} & 10.8 & 8.5 \\
CER-MVS~\cite{cao2024mvsformer++} (K+RT) & \multicolumn{1}{c|}{$\times$} & $\times$ & 21.1 & 24.3 & 11.7 & 47.5 & 4.1 & 71.3 & 6.4 & \multicolumn{1}{c|}{82.1} & 10.8 & 56.3 \\ \midrule
DUSt3R~\cite{wang2024dust3r} & \multicolumn{1}{c|}{$\times$} & med & (3.1) & (71.8) & 3.0 & 76.0 & 3.9 & 68.6 & 3.3 & \multicolumn{1}{c|}{75.1} & 3.3 & 72.9 \\
Pow3R~\cite{jang2025pow3r} & \multicolumn{1}{c|}{$\times$} & med & (3.2) & (68.8) & 3.0 & 74.7 & 3.0 & 74.3 & 3.3 & \multicolumn{1}{c|}{76.6} & 3.1 & 73.6 \\
VGGT~\cite{wang2025vggt} & \multicolumn{1}{c|}{$\times$} & med & (3.7) & (70.0) & 1.7 & 87.2 & 0.9 & 95.4 & 1.7 & \multicolumn{1}{c|}{90.6} & 2.0 & 85.8 \\
\textbf{\ours} & \multicolumn{1}{c|}{$\times$} & med & (3.6) & (72.3) & 1.8 & 87.5 & 1.1 & 93.9 & 1.8 & \multicolumn{1}{c|}{90.0} & 2.1 & 85.9 \\ \midrule
Pow3R~\cite{jang2025pow3r} w/ 100\% (K+RT) & \multicolumn{1}{c|}{$\times$} & med & (3.1) & (71.4) & 2.8 & 77.1 & 1.5 & 91.1 & 3.2 & \multicolumn{1}{c|}{78.2} & 2.7 & 79.5 \\
\textbf{\ours w/ 100\% (K+RT)} & \multicolumn{1}{c|}{$\times$} & med & (3.7) & (72.2) & 1.8 & 87.8 & 1.2 & 93.6 &  1.8 &  \multicolumn{1}{c|}{89.9} & 2.1 & 85.9 \\
\textbf{\ours w/ 30\% D} & \multicolumn{1}{c|}{$\times$} & med & (2.3) & (85.5) & 0.7 & 97.9  & 0.5  & 99.2 & 1.2 & \multicolumn{1}{c|}{93.2} & 1.2 & 94.0 \\
\textbf{\ours w/ 50\% D} & \multicolumn{1}{c|}{$\times$} & med & (2.3) & (85.9) & 0.6 & 98.6  & 0.4  & 99.4 & 1.0 & \multicolumn{1}{c|}{94.4} & 1.1 & 94.6 \\
\textbf{\ours w/ 70\% D} & \multicolumn{1}{c|}{$\times$} & med & (2.3) & (\underline{86.0}) & 0.6 & 98.7  & 0.4  & 99.4 & 0.9 & \multicolumn{1}{c|}{95.5} & 1.1 & \underline{94.9} \\
\textbf{\ours w/ 100\% D} & \multicolumn{1}{c|}{$\times$} & med & (\underline{2.3}) & (85.6) & \underline{0.5} & \underline{98.7}  & \underline{0.3}  & \textbf{99.5} & \underline{0.9} & \multicolumn{1}{c|}{\underline{95.5}} & \underline{1.0} & 94.8 \\
\textbf{\ours w/ 100\% (K+RT+D)} & \multicolumn{1}{c|}{$\times$} & med & (\textbf{2.2}) & (\textbf{86.7}) & \textbf{0.5} & \textbf{98.7} & \textbf{0.3} & \underline{99.4} & \textbf{0.9} & \multicolumn{1}{c|}{\textbf{95.6}} & \textbf{1.0} & \textbf{95.1} \\ \bottomrule
\end{tabular}}
\end{table*}

\subsection{More Results of 3D Reconstrunction}\label{more3dreconstruction}

In this subsection, we present the complete 3D reconstruction results. 
Table~\ref{tab:full7scenereconstruct} and \ref{tab:fullnrgbreconstruct} report the benchmark results on the 7-Scenes~\cite{shotton2013scene} and NRGB~\cite{azinovic2022neural} datasets, respectively, including the effects of incorporating different types and percentages of auxiliary information.

\begin{table*}[t]
    \centering
    \begin{minipage}{0.48\linewidth}
        \centering
        \caption{\textbf{3D reconstruction on the 7-scenes~\cite{shotton2013scene} datasets.} K”, “RT”, and “D” denote intrinsic, relative pose, and depth information, respectively. The best and second best results in each category are \textbf{bold} and \underline{underlined} respectively.}
        \label{tab:full7scenereconstruct}
        \centering
        \footnotesize
        \resizebox{\linewidth}{!}{
        \begin{tabular}{@{}l|cc|cc|cc@{}}
        \toprule
        \multicolumn{1}{c|}{\multirow{2}{*}{\textbf{Method}}} & \multicolumn{2}{c|}{\textbf{Acc$\downarrow$}} & \multicolumn{2}{c|}{\textbf{Comp$\downarrow$}} & \multicolumn{2}{c|}{\textbf{NC$\uparrow$}}  \\ \cmidrule(lr){2-7}
         & \textbf{Mean} & \textbf{Med.} & \textbf{Mean} & \textbf{Med.} & \textbf{Mean} & \textbf{Med.}   \\ \midrule
        \textbf{VGGT~\cite{wang2025vggt}} & \textbf{0.087} & \underline{0.039} & \textbf{0.091} & \underline{0.039} & \textbf{0.787} & \textbf{0.890}  \\
        Fast3R~\cite{yang2025fast3r} & 0.164 & 0.108 & 0.163 & 0.080 & 0.686 & 0.775  \\
        DUSt3R-GA~\cite{wang2024dust3r} & 0.146 & 0.077 & 0.181 & 0.067 & 0.736 & 0.839  \\
        MASt3R-GA~\cite{leroy2024mast3r} & 0.185 & 0.081 & 0.180 & 0.069 & 0.701 & 0.792  \\
        MonST3R-GA~\cite{zhang2024monst3r} & 0.248 & 0.185 & 0.266 & 0.167 & 0.672 & 0.759  \\
        Spann3R~\cite{wang20243d} & 0.298 & 0.226 & 0.205 & 0.112 & 0.650 & 0.730  \\
        SLAM3R~\cite{liu2025slam3r} & 0.287 & 0.155 & 0.226 & 0.066 & 0.644 & 0.720  \\
        CUT3R~\cite{wang2025cut3r} & 0.126 & 0.047 & 0.154 & 0.031 & 0.727 & 0.834  \\
        \textbf{\ours} & \underline{0.104} & \textbf{0.037} & \underline{0.112} & \textbf{0.031} & \underline{0.763} & \underline{0.875}  \\ \midrule[0.8pt]
        \textbf{\shortstack[l]{\ours \\ w/ 30\% D} }  & 0.098 & 0.034 & 0.110 & 0.028 & 0.778 & 0.889  \\
        \textbf{\shortstack[l]{\ours \\ w/ 50\% D} }  & 0.103 & 0.033 & 0.106 & 0.028 & 0.781 & 0.890 \\
        \textbf{\shortstack[l]{\ours \\ w/ 70\% D} }  & 0.094 & 0.045 & 0.094 & 0.033 & 0.785 & 0.885 \\
        \textbf{\shortstack[l]{\ours \\ w/ 100\% D} }  & 0.085 & 0.034 & 0.085 & 0.027 & \underline{0.789} & \underline{0.894}  \\ \midrule
        \textbf{\shortstack[l]{\ours \\ w/ 30\% (K+RT)}} & 0.100 & 0.034 & 0.107 & 0.027 & 0.761 & 0.875  \\
        \textbf{\shortstack[l]{\ours \\ w/ 50\% (K+RT)}} & 0.101 & 0.035 & 0.108 & 0.027 & 0.761 & 0.875  \\
        \textbf{\shortstack[l]{\ours \\ w/ 70\% (K+RT)}} & 0.089 & 0.022 & 0.096 & 0.022 & 0.771 & 0.886  \\
        \textbf{\shortstack[l]{\ours \\ w/ 100\% (K+RT)}} & \underline{0.037} & \underline{0.017} & \underline{0.049} & \underline{0.019} &  0.778 & 0.893  \\ \midrule
        \textbf{\shortstack[l]{\ours \\ w/ 100\% (K+RT+D)}} & \textbf{0.036} & \textbf{0.017} & \textbf{0.036} & \textbf{0.017} &  \textbf{0.810} & \textbf{0.912}  \\
        \bottomrule
        \end{tabular}}
    \end{minipage}\hfill
    \begin{minipage}{0.48\linewidth}
        \centering
        \caption{\textbf{3D reconstruction on the NRGB~\cite{azinovic2022neural} datasets.} K”, “RT”, and “D” denote intrinsic, relative pose, and depth information, respectively. The best and second best results in each category are \textbf{bold} and \underline{underlined} respectively.}
        \label{tab:fullnrgbreconstruct}
        \centering
        \footnotesize
        \resizebox{\linewidth}{!}{
        \begin{tabular}{@{}l|cc|cc|cc@{}}
        \toprule
        \multicolumn{1}{c|}{\multirow{2}{*}{\textbf{Method}}} & \multicolumn{2}{c|}{\textbf{Acc$\downarrow$}} & \multicolumn{2}{c|}{\textbf{Comp$\downarrow$}} & \multicolumn{2}{c|}{\textbf{NC$\uparrow$}}  \\ \cmidrule(lr){2-7}
         & \textbf{Mean} & \textbf{Med.} & \textbf{Mean} & \textbf{Med.} & \textbf{Mean} & \textbf{Med.}   \\ \midrule
        \textbf{VGGT~\cite{wang2025vggt}} & \textbf{0.087} & \underline{0.039} & \textbf{0.091} & \underline{0.039} & \textbf{0.787} & \textbf{0.890}  \\
        Fast3R~\cite{yang2025fast3r} & 0.164 & 0.108 & 0.163 & 0.080 & 0.686 & 0.775  \\
        DUSt3R-GA~\cite{wang2024dust3r} & 0.146 & 0.077 & 0.181 & 0.067 & 0.736 & 0.839  \\
        MASt3R-GA~\cite{leroy2024mast3r} & 0.185 & 0.081 & 0.180 & 0.069 & 0.701 & 0.792  \\
        MonST3R-GA~\cite{zhang2024monst3r} & 0.248 & 0.185 & 0.266 & 0.167 & 0.672 & 0.759  \\
        Spann3R~\cite{wang20243d} & 0.298 & 0.226 & 0.205 & 0.112 & 0.650 & 0.730  \\
        SLAM3R~\cite{liu2025slam3r} & 0.287 & 0.155 & 0.226 & 0.066 & 0.644 & 0.720  \\
        CUT3R~\cite{wang2025cut3r} & 0.126 & 0.047 & 0.154 & 0.031 & 0.727 & 0.834  \\
        \textbf{\ours} & \underline{0.104} & \textbf{0.037} & \underline{0.112} & \textbf{0.031} & \underline{0.763} & \underline{0.875}  \\ \midrule[0.8pt]
        \textbf{\shortstack[l]{\ours \\ w/ 30\% D} }  & 0.098 & 0.034 & 0.110 & 0.028 & 0.778 & 0.889  \\
        \textbf{\shortstack[l]{\ours \\ w/ 50\% D} }  & 0.103 & 0.033 & 0.106 & 0.028 & 0.781 & 0.890 \\
        \textbf{\shortstack[l]{\ours \\ w/ 70\% D} }  & 0.094 & 0.045 & 0.094 & 0.033 & 0.785 & 0.885 \\
        \textbf{\shortstack[l]{\ours \\ w/ 100\% D} }  & 0.085 & 0.034 & 0.085 & 0.027 & \underline{0.789} & \underline{0.894}  \\ \midrule
        \textbf{\shortstack[l]{\ours \\ w/ 30\% (K+RT)}} & 0.100 & 0.034 & 0.107 & 0.027 & 0.761 & 0.875  \\
        \textbf{\shortstack[l]{\ours \\ w/ 50\% (K+RT)}} & 0.101 & 0.035 & 0.108 & 0.027 & 0.761 & 0.875  \\
        \textbf{\shortstack[l]{\ours \\ w/ 70\% (K+RT)}} & 0.089 & 0.022 & 0.096 & 0.022 & 0.771 & 0.886  \\
        \textbf{\shortstack[l]{\ours \\ w/ 100\% (K+RT)}} & \underline{0.037} & \underline{0.017} & \underline{0.049} & \underline{0.019} &  0.778 & 0.893  \\ \midrule
        \textbf{\shortstack[l]{\ours \\ w/ 100\% (K+RT+D)}} & \textbf{0.036} & \textbf{0.017} & \textbf{0.036} & \textbf{0.017} &  \textbf{0.810} & \textbf{0.912}  \\
        \bottomrule
        \end{tabular}}
    \end{minipage}
\end{table*}

\subsection{More Visualizations}
\begin{figure*}[t]
\begin{center}
\includegraphics[width=1\linewidth]{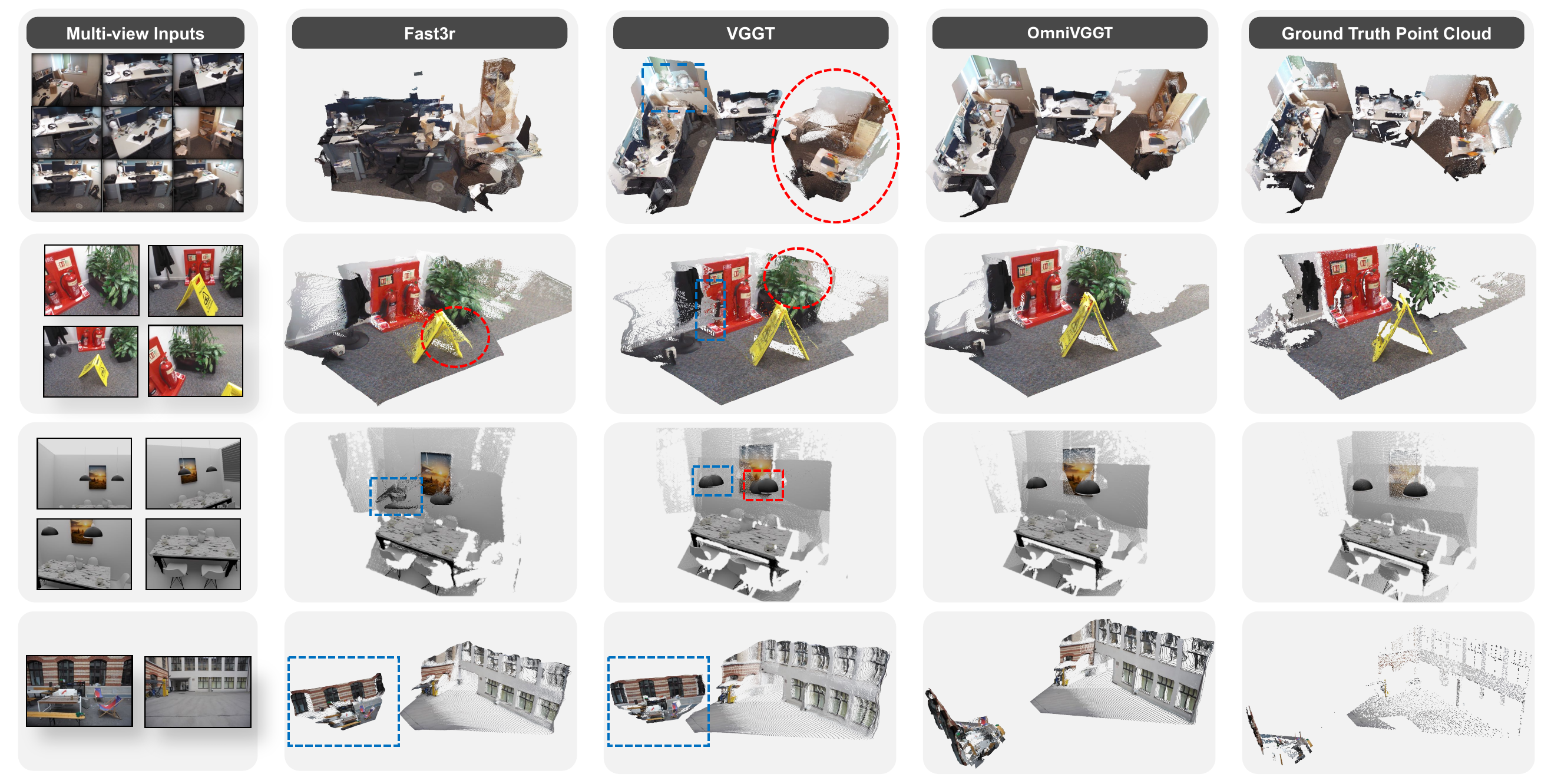}
\caption{\textbf{Visual Comparisons on 7-Scenes~\cite{shotton2013scene}, NRGBD~\cite{azinovic2022neural}, and ETH3D~\cite{schops2017multi} datasets.}}
\label{fig:visualization2}
\end{center}
\end{figure*}

Fig.~\ref{fig:visualization2} presents a visual comparison between \ours and other methods on the 7-Scenes\cite{shotton2013scene}, NRGBD\cite{silberman2012indoor}, and ETH3D~\cite{schops2017multi} datasets. 
With the assistance of additional modality inputs (e.g., camera poses), \ours maintains accurate spatial relationships even in extreme cases—such as when two images have no overlap at all. 
Fig.~\ref{fig:inthewild} further demonstrates the strong generalization ability of \ours on in-the-wild data, where it achieves impressive results across both synthetic-engine-rendered and AI-generated images.
Fig.~\ref{fig:visualization3} illustrates the reconstruction performance on image pairs that contain limited or no overlap between views.
We also show the rollout examples of our method on the CALVIN benchmark in Fig.~\ref{fig:calvin_rollout}.

\begin{figure*}[t]
\begin{center}
\includegraphics[width=1\linewidth]{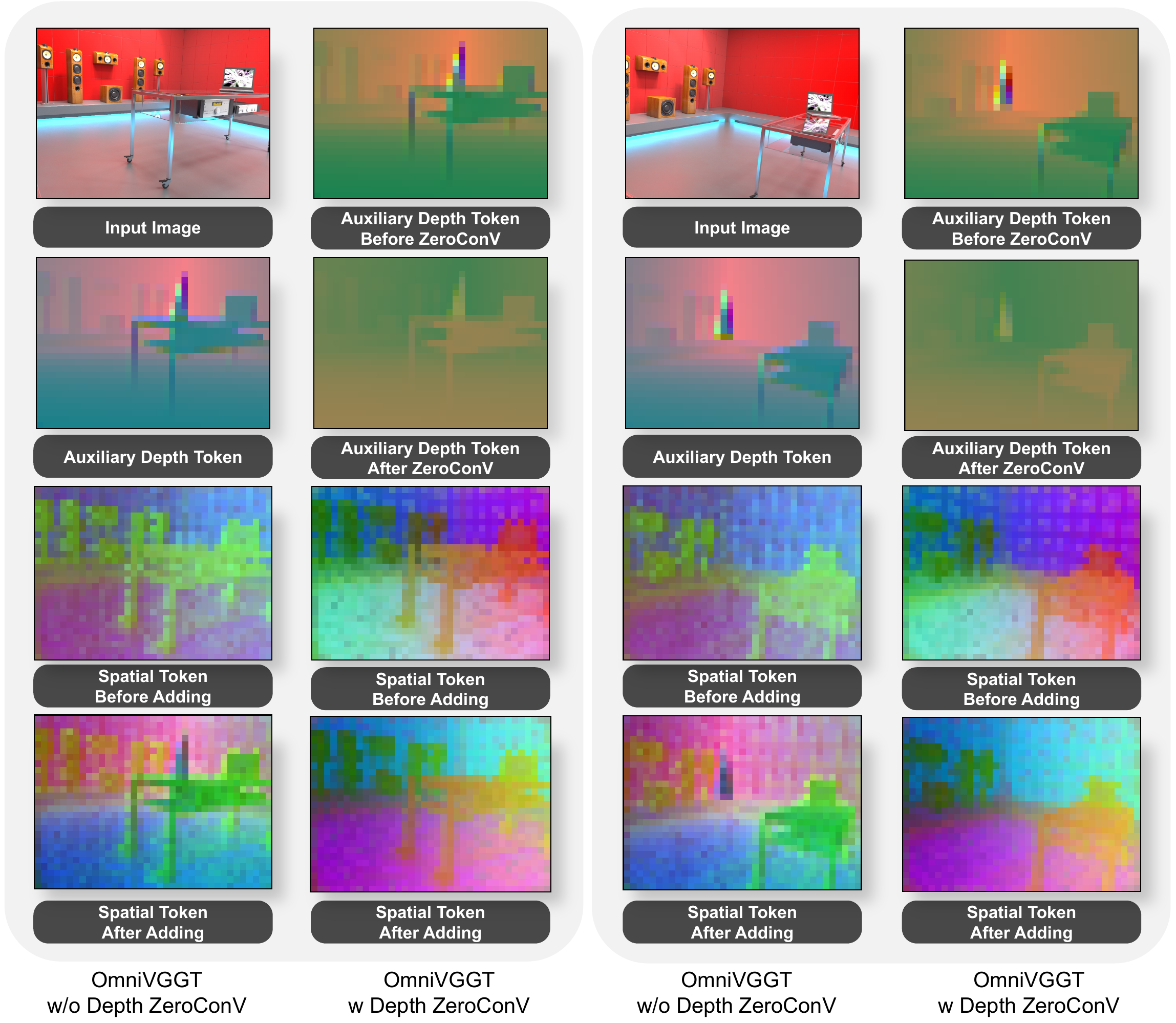}
\caption{\textbf{Feature Map Visualization Comparison Between \ours and \ours w/ Depth ZeroConV. } \textbf{Left Column}: \ours without applying the ZeroConV to the depth adapter. From top to bottom: the input image, the feature map of the auxiliary depth token after passing through the depth encoder; the spatial token feature map obtained from the image after the DINO encoder; and the spatial token feature map after adding the auxiliary depth tokens. \textbf{Right Colum}: \ours with the ZeroConV applied to the auxiliary depth information. From top to bottom: the auxiliary depth token feature map before and after the ZeroConV; and the spatial token feature maps before and after the addition of auxiliary depth tokens.}
\label{fig:alb_visualization}
\end{center}
\vspace{-5mm}
\end{figure*}

\begin{figure*}[t]
    \centering
    \begin{minipage}{1\linewidth}
        \centering
        \includegraphics[width=0.95\linewidth]{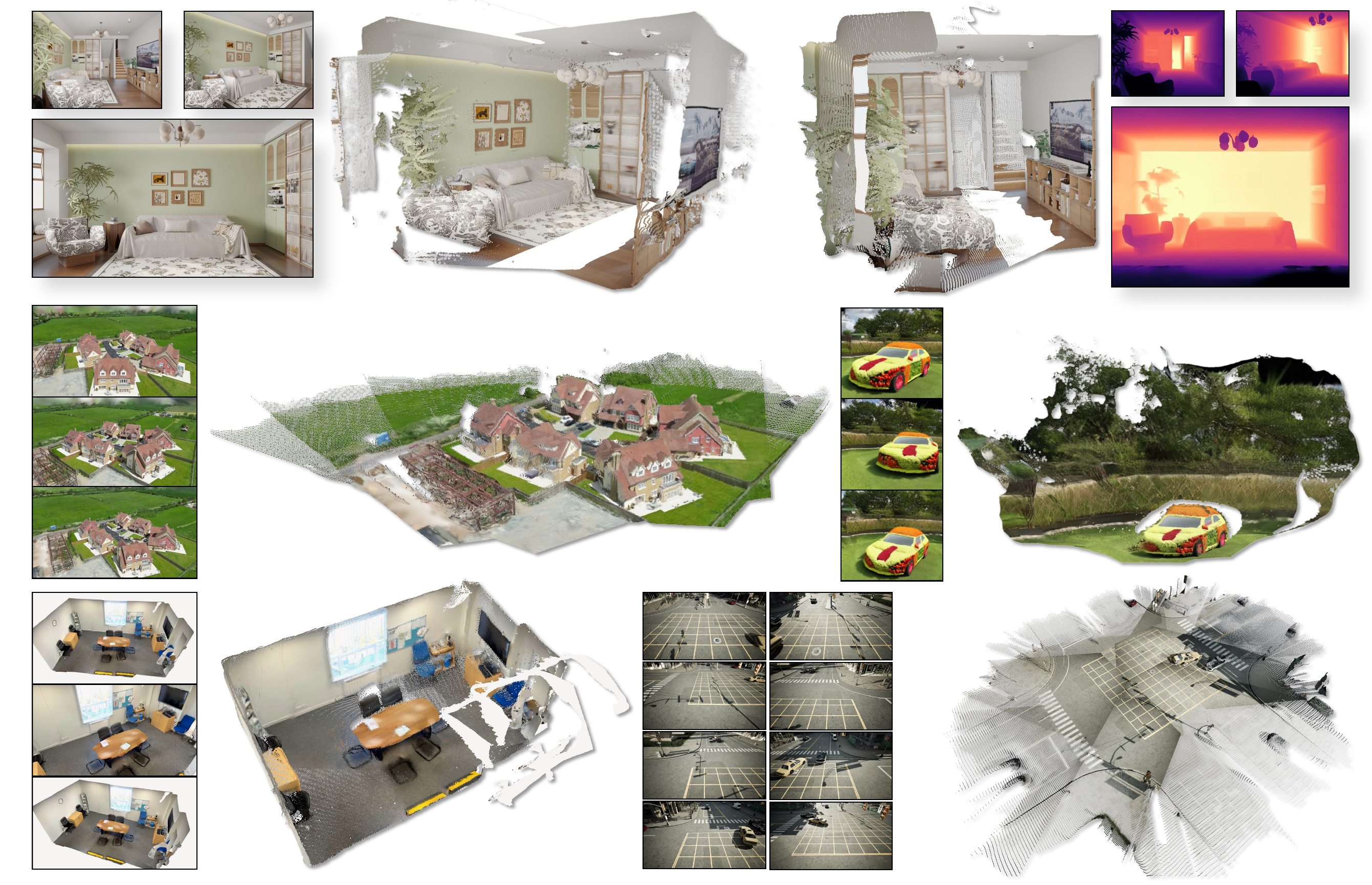}
        \caption{\textbf{Feed-Forward 3D Point Map by \ours with In-The-Wild Inputs.}}
        \label{fig:inthewild}
    \end{minipage}
    \hfill
    \begin{minipage}{1\linewidth}
        \centering
        \includegraphics[width=0.95\linewidth]{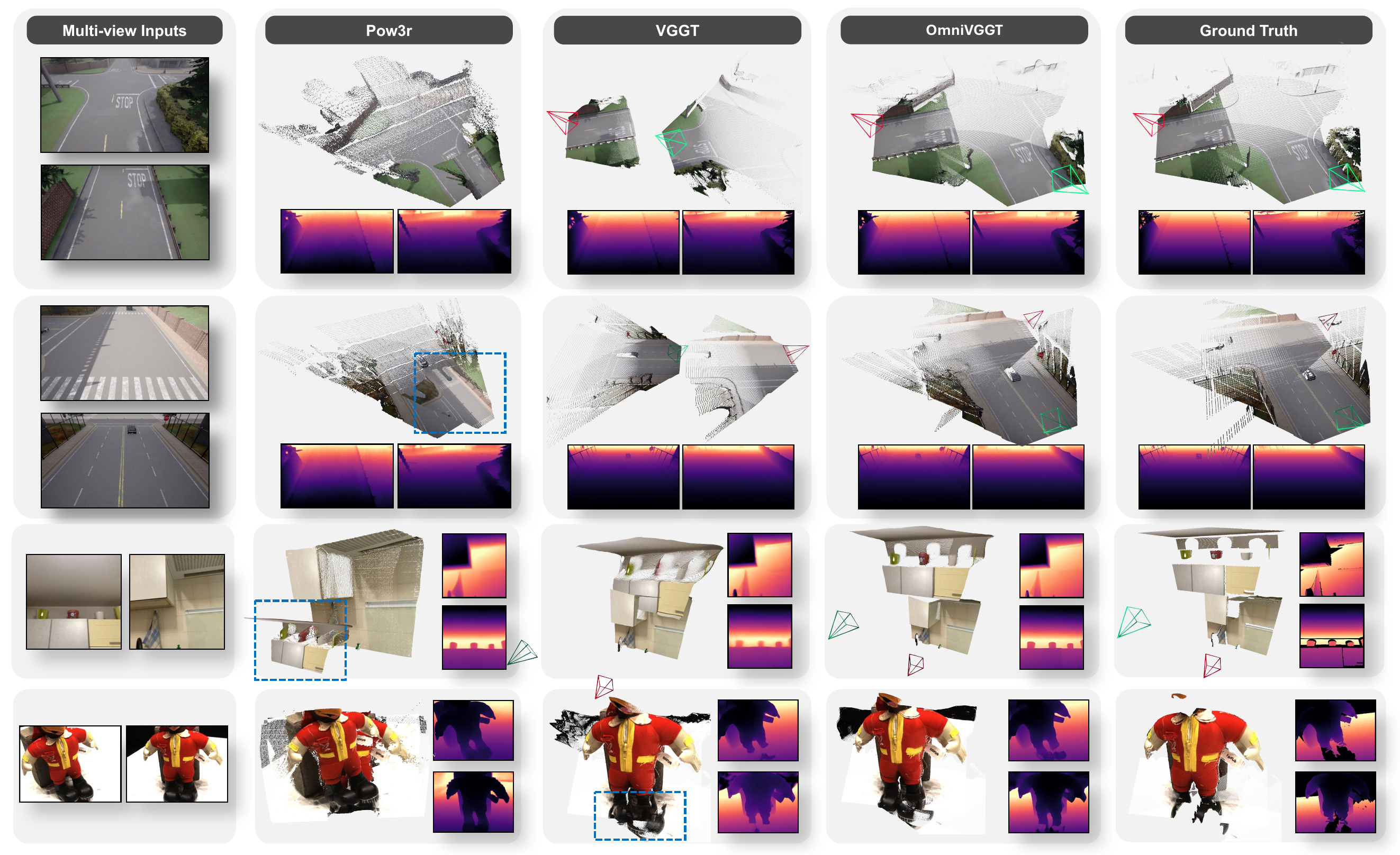}
        \caption{\textbf{More Visualizations on Image Pairs Input.}}
        \label{fig:visualization3}
    \end{minipage}
\end{figure*}

\begin{table*}[t]
\caption{\textbf{Ablation of GeoAdapter architectures.} We compare different GeoAdapter designs on the Sintel~\cite{butler2012naturalistic} dataset. 
The `Replace' refers to directly substituting the original camera tokens with auxiliary camera tokens, whereas `One-Layer GeoAdapter' represents injecting auxiliary camera tokens only once before the encoder. `Depth ZeroConV' denotes using ZeroConV for depth injection.}
\label{tab:fullablation}
\vspace{-1mm}
\centering
\footnotesize
\resizebox{0.7\linewidth}{!}{
\begin{tabular}{@{}l|cc|ccc@{}}
\toprule
\multicolumn{1}{c|}{\multirow{2}{*}{\textbf{Architecture}}} & 
\multicolumn{2}{c|}{\textbf{Depth}} & 
\multicolumn{3}{c}{\textbf{Camera}} \\ 
\cmidrule(l){2-6}
 & \textbf{Abs Rel$\downarrow$} & \textbf{$\delta<1.25\uparrow$} & 
 RRA@$5^{\circ}\uparrow$ & RTA@$5^{\circ}\uparrow$ & AUC@$30^{\circ}\uparrow$ \\ 
\midrule

\multicolumn{6}{l}{\textit{w/o (K+RT+D) Auxiliary Information}} \\ 
\cmidrule(l){1-6}

\textbf{(a) Replace} & 0.845 & 64.74 & 93.40 & 30.88 & 64.74 \\
\textbf{(b) One-Layer Adapter} & 0.604 & 68.74 & 96.78 & 44.92 & 68.74 \\
\textbf{(c) Depth ZeroConV} & 0.569 & 70.71 & 96.44 & 51.86 & 69.70 \\
\textbf{(d) \ours} & \textbf{0.558} & \textbf{71.46} & 96.15 & \textbf{54.01} & \textbf{70.83} \\ 
\midrule

\multicolumn{6}{l}{\textit{w/ (K+RT) Auxiliary Information}} \\ 
\cmidrule(l){1-6}

\textbf{(a) Replace} & 0.842 & 65.11 & 96.98 & 55.66 & 76.28 \\
\textbf{(b) One-Layer Adapter} & 0.563 & 70.16 & 97.54 & 58.84 & 77.00 \\
\textbf{(c) Depth ZeroConV} & 0.569 & 70.71 & 99.65 & 69.29 & 83.18 \\
\textbf{(d) \ours} & \textbf{0.553} & \textbf{72.36} & \textbf{99.97} & \textbf{75.83} & \textbf{85.35} \\
\midrule

\multicolumn{6}{l}{\textit{w/ Depth Auxiliary Information}} \\ 
\cmidrule(l){1-6}

\textbf{(a) Replace} & 0.670 & 82.96 & 94.69 & 52.04 & 71.61 \\
\textbf{(b) One-Layer Adapter} & 0.107 & 85.91 & 95.97 & 52.34 & 74.66 \\
\textbf{(c) Depth ZeroConV} & 0.570 & 71.02 & 95.45 & 56.10 & 72.93 \\
\textbf{(d) \ours} & \textbf{0.106} & \textbf{85.95} & \textbf{96.93} & \textbf{59.73} & \textbf{77.16} \\
\midrule

\multicolumn{6}{l}{\textit{w/ (K+RT+D) Auxiliary Information}} \\ 
\cmidrule(l){1-6}

\textbf{(a) Replace} & 0.655 & 82.96 & 97.08 & 57.61 & 77.83 \\
\textbf{(b) One-Layer Adapter} & 0.133 & 85.65 & 99.97 & 60.89 & 81.66 \\
\textbf{(c) Depth ZeroConV} & 0.505 & 71.11 & 99.72 & 71.66 & 84.12 \\
\textbf{(d) \ours} & \textbf{0.106} & \textbf{85.95} & \textbf{99.97} & \textbf{76.33} & \textbf{85.99} \\

\bottomrule
\end{tabular}}
\end{table*}

\subsection{Architecture Ablation}
In this subsection, we investigate the impact of different GeoAdapter strategies on model performance. 
We randomly select 100 batches from the Sintel~\cite{butler2012naturalistic} dataset, with 10 images per scene for inference, and evaluate the related metrics for depth and camera estimation. 
We trained the following three additional versions under the same training parameters and dataset.
We experiment with three strategies: (a) \textit{Replace}, where the original camera tokens are directly replaced by auxiliary camera tokens in each AA block, (b) \textit{One-Layer Adapter}, where auxiliary camera tokens are only injected once before the AA blocks,  and (c) \textit{Depth ZeroConV}, where a ZeroConV layer is used for depth information injection.
Table~\ref{tab:fullablation} reports the full results of our ablation experiments.

In addition, Fig.~\ref{fig:alb_visualization} illustrates the feature maps of the spatial tokens and auxiliary depth tokens in \ours, with and without the use of the depth ZeroConV. 
We use Principal Component Analysis (PCA) to reduce the high-dimensional token embeddings to three dimensions and save them as RGB images.
We observe that when the ZeroConv layer is applied to depth injection, the meaningful information within the auxiliary depth features (e.g., edges and background structures) is largely suppressed, causing the network to treat the injected depth as noise. 
In this case, the feature representations of OmniVGGT with and without auxiliary depth inputs remain nearly identical, and the resulting performance shows almost no difference.
Therefore, we chose not to include the depth ZeroConV in our final network design.

\begin{figure*}[t]
\begin{center}
    \includegraphics[width=\linewidth,height=\textheight,keepaspectratio]{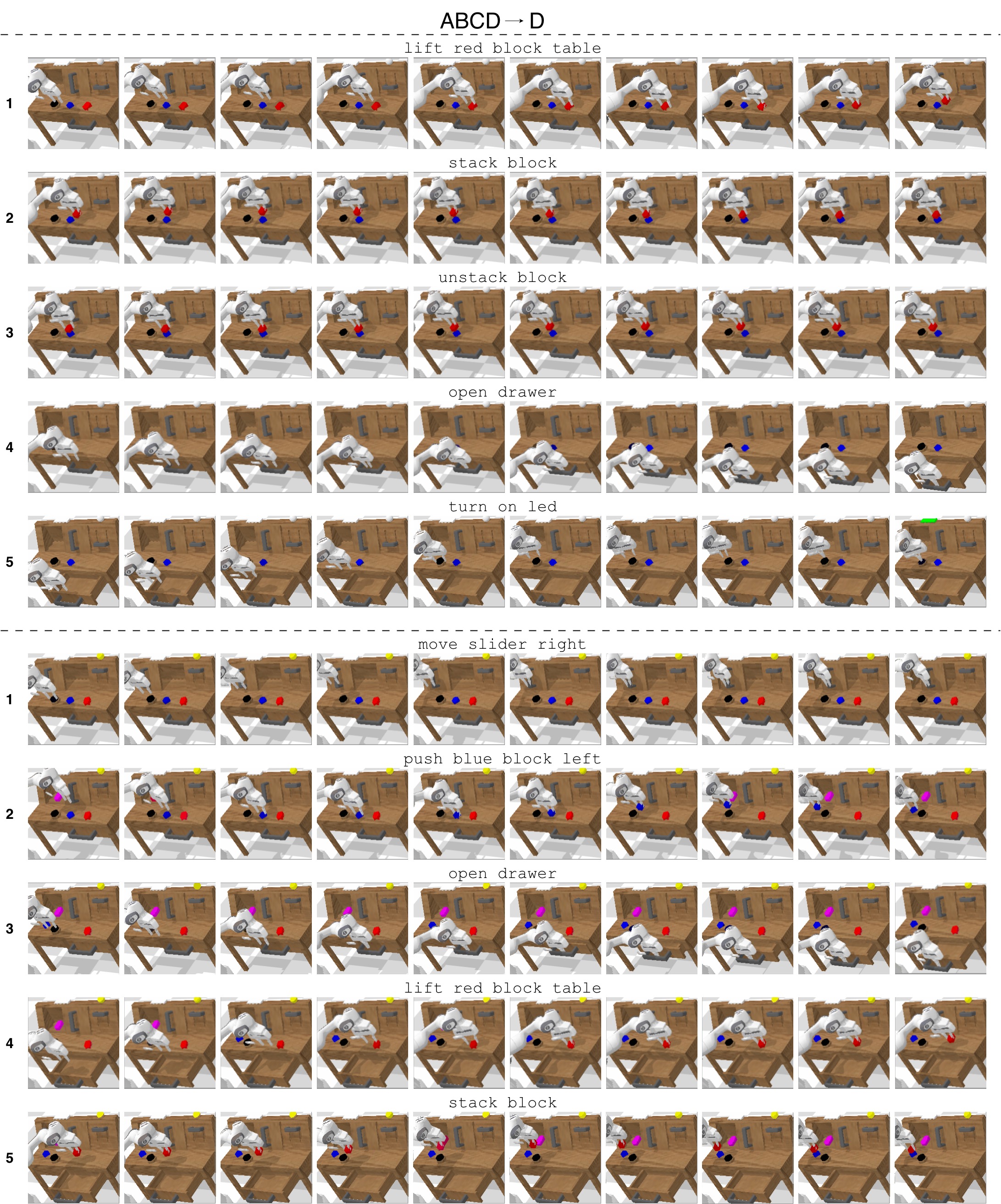}
    \captionof{figure}{\textbf{Ours Rollouts on the ABCD$\rightarrow$D split of the CALVIN benchmark.}}
    \label{fig:calvin_rollout}
\end{center}
\end{figure*}


\end{document}